\documentclass{article}

\usepackage{microtype}
\usepackage{graphicx}
\usepackage{subfigure}
\usepackage{booktabs} 

\usepackage{hyperref}

\usepackage[accepted]{icml2023}

\usepackage{amsmath}
\usepackage{amssymb}
\usepackage{mathtools}
\usepackage{amsthm}

\usepackage[capitalize,noabbrev]{cleveref}

\theoremstyle{plain}
\newtheorem{theorem}{Theorem}[section]

\theoremstyle{definition}
\newtheorem{definition}[theorem]{Definition}

\theoremstyle{remark}

\usepackage[textsize=tiny]{todonotes}

\usepackage{amsmath,amsfonts,bm,amsthm}

\def\eqref#1{equation~\ref{#1}}

\def\1{\bm{1}}

\DeclareMathAlphabet{\mathsfit}{\encodingdefault}{\sfdefault}{m}{sl}
\SetMathAlphabet{\mathsfit}{bold}{\encodingdefault}{\sfdefault}{bx}{n}

\newcommand{\E}{\mathbb{E}}

\newcommand{\R}{\mathbb{R}}

\DeclareMathOperator*{\argmax}{arg\,max}
\DeclareMathOperator*{\argmin}{arg\,min}

\newtheorem{innercustomgeneric}{\customgenericname}
\providecommand{\customgenericname}{}
\newcommand{\newcustomtheorem}[2]{%
  \newenvironment{#1}[1]
  {%
   \renewcommand\customgenericname{#2}%
   \renewcommand\theinnercustomgeneric{##1}%
   \innercustomgeneric
  }
  {\endinnercustomgeneric}
}
\newcustomtheorem{manualtheorem}{Theorem}

\def\Pr{\mathbf{Pr}}

\newcustomtheorem{manuallemma}{Lemma}
\newcustomtheorem{manualproposition}{Proposition}
\newcustomtheorem{manualassumption}{Assumption}

\newcommand\omicron{o}

\icmltitlerunning{Monte-Carlo tree search with uncertainty propagation via optimal transport}

\begin{document}

\twocolumn[
\icmltitle{Monte-Carlo tree search with uncertainty propagation via optimal transport}




\begin{icmlauthorlist}
\icmlauthor{Tuan Dam}{1}
\icmlauthor{Pascal Stenger}{4}
\icmlauthor{Lukas Schneider}{6}
\icmlauthor{Joni Pajarinen}{5}
\icmlauthor{Carlo D'Eramo}{2,3}
\icmlauthor{Odalric-Ambrym Maillard}{1}
\end{icmlauthorlist}

\icmlaffiliation{1}{Univ. Lille, Inria, CNRS, Centrale Lille, UMR 9189-CRIStAL, F-59000 Lille, France}
\icmlaffiliation{2}{Center for Artificial Intelligence and Data Science, University of W\"{u}rzburg, Germany}
\icmlaffiliation{3}{Hessian.ai, Germany}
\icmlaffiliation{4}{Department of Computer Science, Technical University of Darmstadt, Germany}
\icmlaffiliation{5}{Department of Electrical Engineering and Automation, Aalto University, Finland}
\icmlaffiliation{6}{ETHZ - ETH Zurich, Switzerland}

\icmlcorrespondingauthor{Tuan Dam}{tuan.dam@inria.fr}

\icmlkeywords{Machine Learning, ICML}

\vskip 0.3in
]



\printAffiliationsAndNotice{}  

\begin{abstract}
 This paper introduces a novel backup strategy for Monte-Carlo Tree Search (MCTS) designed for highly stochastic and partially observable Markov decision processes. We adopt a probabilistic approach, modeling both value and action-value nodes as Gaussian distributions. We introduce a novel backup operator that computes value nodes as the Wasserstein barycenter of their action-value children nodes; thus, propagating the uncertainty of the estimate across the tree to the root node. We study our novel backup operator when using a novel combination of $L^1$-Wasserstein barycenter with $\alpha$-divergence, by drawing a notable connection to the generalized mean backup operator. We complement our probabilistic backup operator with two sampling strategies, based on optimistic selection and Thompson sampling, obtaining our Wasserstein MCTS algorithm. We provide theoretical guarantees of asymptotic convergence to the optimal policy, and an empirical evaluation on several stochastic and partially observable environments, where our approach outperforms well-known related baselines.

\end{abstract}

\section{Introduction}
Monte-Carlo tree search (MCTS) is a search algorithm that combines elements of planning and reinforcement learning (RL)~\cite{sutton-barto:book} to solve decision-making problems. Recently, the coupling of MCTS with deep learning techniques for value estimation enabled solving complex problems with high branching factor, considered unsolvable only a few years ago~\cite{alphago,alphazero,schrittwieser2020mastering}. These impressive results have been achieved mostly in deterministic problems. However, in highly stochastic environments, possibly with partial observability, value estimation becomes largely more challenging. High uncertainty can cause inaccurate estimates of value functions, which propagate across the search tree to the root node, leading to suboptimal action selection and poor performance. In this work, we introduce a novel MCTS algorithm to account for the uncertainty due to stochasticity and partial observability. Our method is a probabilistic approach that handles the uncertainty of value nodes by modeling them as Gaussian distributions, and adopts a novel backup operator for MCTS which leverages the notion of Wasserstein barycenters~\cite{wsteinbarry} to propagate values and their uncertainty across the tree, to the root node. For this reason, we call our new method \textit{Wasserstein MCTS}. We introduce the use of $\alpha$-divergence~\cite{fdiv} as a measure of distance to compute $L^1$-Wasserstein barycenters, discovering that we can retrieve the generalized mean backup operator introduced in~\citet{dam2019generalized}. This operator adopts a power mean backup which unifies the commonly used average backup operator and maximum backup operator through a tunable hyperparameter, aiming at tackling the value overestimation issue often affecting RL methods ~\cite{hasselt2010double}. Our approach is a probabilistic variant that unifies regular MCTS methods and enables us to account for the uncertainty in the estimate of value nodes, similar to the approach proposed in ~\cite{metelli2019propagating} for uncertainty propagation in temporal-difference learning. We couple our backup operator with suitable exploration strategies; namely, we investigate using posterior sampling and Thompson sampling ~\cite{Thompson}. We further investigate the latter exploration strategy to prove the asymptotic convergence of our Wasserstein MCTS algorithm to the optimal policy, by analyzing the multi-armed bandit problem from the root node of the tree. We complement our theoretical findings, with an empirical evaluation of our Wasserstein MCTS algorithm in problems where high stochasticity in the dynamics and partial observability pose a serious challenge, evincing tangible advantages of our approach w.r.t. state-of-the-art MCTS standard and probabilistic methods.

Our \textit{contribution} is threefold: 1) we introduce a novel MCTS algorithm to tackle highly stochastic and partially observable problems, that models value nodes as Gaussian distributions to account for uncertainty in their estimate, and applies a backup operator based on Wasserstein barycenters, to propagate the uncertainty across the tree. We also present a particle filter approach that can be applied to any distribution, and we believe that this approach is not limited to the Gaussian assumption; 2) we are the first to apply Wasserstein Barycenter to backpropagate value functions in MCTS and establish the connection to the power mean backup operator used in Power-UCT; 3) we prove that our Wasserstein MCTS algorithm, coupled with Thompson sampling, enjoys advantageous asymptotic convergence to the optimal policy, as previously observed in nonprobabilistic MCTS methods~\cite{dam2019generalized}.
\vspace{-0.5cm}
\section{Related work}\label{sec_related_work}
\citet{metelli2019propagating} introduces the use of $L^2$-Wasserstein barycenters in combination with Euclidean distance to propagate the uncertainty of the action-value estimate in a temporal-difference learning manner. The resulting Wasserstein $Q$-Learning algorithm exhibits remarkable advantages over alternative methods in stochastic environments.

In MCTS, several papers provide Bayesian approaches to account for uncertainty in the value function estimation. \citet{tesauro2012bayesian} propose to represent the value function at each node in the tree as a Gaussian distribution and introduce an alternative exploration strategy for action selection in the tree, that uses the standard deviation of a Gaussian for better estimation of the exploration constant. \citet{bai2013bayesian} model the value function at each node in the tree using a Dirichlet-NormalGamma distribution coupled with Thompson sampling exploration, proving convergence to the selection of the optimum and showing the advantages in highly stochastic environments. \citet{bai2014thompson} extend this method to partially observable problems. 
We also use Thompson sampling for action selection, but, opposite to the mentioned approaches, we leverage the original idea of propagating the uncertainty of the action-value estimates across the tree, and for the introduction of $L^1$-Wasserstein barycenters in combination with the $\alpha$-divergence, to additionally handle highly stochastic environments.

Generalized mean backup operators are extensively studied in \cite{dam2019generalized} to tackle the underestimation/overestimation problems of average and maximum backup operators in UCT~\cite{coulom2007efficient}. As previously mentioned, our Wasserstein MCTS algorithm, thanks to the original use of $L^1$-Wasserstein barycenters with $\alpha$-divergence, can be shown to result in the same generalized mean backup operator proposed in \citet{dam2019generalized}; thus, in combination with the probabilistic perspective adopted by Wasserstein MCTS, make it an effective solution to tackle highly stochastic problems.

In multi-armed bandits (MAB), two most popular policy selection strategies to solve the exploration-exploitation dilemma are optimism in the face of uncertainty~\cite{auer2002finite} and Thompson sampling~\cite{thompson1933likelihood}. We consider these two sampling methods, coupled with our novel probabilistic perspective on uncertainty propagation in MCTS, to select actions and expand the search tree.
\vspace{-0.3cm}
\section{Background}\label{sec_background}
\subsection{Markov decision process}\label{sec_mdp}
We consider an agent interacting in an environment modeled as a infinite-horizon discounted Markov decision process~(MDP) $\mathcal{M} = \langle \mathcal{S}, \mathcal{A}, \mathcal{R}, \mathcal{P}, \gamma \rangle$, where $\mathcal{S}$ is the state space, $\mathcal{A}$ is the finite discrete action space, $\mathcal{R}: \mathcal{S} \times \mathcal{A} \times \mathcal{S} \to \mathbb{R}$ is the reward function, $\mathcal{P}: \mathcal{S} \times \mathcal{A} \to \mathcal{S}$ is the transition kernel, and $\gamma \in [0, 1)$ is the discount factor. A policy $\pi \in \Pi: \mathcal{S} \to \mathcal{A}$ is the probability distribution of executing actions conditioned on states. A policy induces an action-value function $Q^\pi \triangleq \mathbb{E} \left[\sum_{k=0}^\infty \gamma^k r_{i+k+1} | s_i = s, a_i = a, \pi \right]$, which is the expected cumulative discounted reward collected executing action $a$ in state $s$, following policy $\pi$ thereafter,  where $r_{i+1}$ is the reward obtained after the $i$-th transition. The goal of agents is to find the optimal policy that maximizes the action-value function, in other words, the one that satisfies the Bellman equation~\cite{bellman1954theory} $Q^*(s,a) \triangleq \int_{\mathcal{S}} \mathcal{P}(s'|s,a)\left[ \mathcal{R}(s,a,s') + \gamma \max_{a'}Q^*(s',a') \right] ds', \forall s \in \mathcal{S}, a \in \mathcal{A}$. Given an optimal action-value function, we can compute the optimal value function as $V^{*}(s) \triangleq \max_{a \in \mathcal{A}} Q^*(s,a), \forall s \in \mathcal{S}$.
\subsection{Monte-Carlo tree search}\label{sec_mcts}
Monte-Carlo tree search~(MCTS) combines Monte-Carlo sampling, tree search, and exploration strategies inspired by multi-armed bandits~\cite{ucb}, to solve MDPs. MCTS builds a search tree where visited states are modeled as nodes, and actions executed in each state as edges. MCTS consists of four steps:
\underline{Selection:} a \textit{tree-policy} is executed to navigate the tree from the root node until the leaf node is reached;\underline{Expansion:} the reached node is expanded according to the tree policy; \underline{Simulation:} execute a rollout, e.g. Monte-Carlo simulation, from the visited child of the current node to the end of the episode, to estimate the value of the newly added node. Another way is to estimate this value from a pretrained neural network~\cite{alphago};\underline{Backup:} use collected reward to update the action-values $Q(\cdot)$ along the visited trajectory.

In the next section, we present the baseline method Upper Confidence bound for Trees (UCT)~\cite{kocsis2006improved}.
\vspace{-0.2cm}
\subsection{Upper Confidence bound for trees}\label{sec_uct}
The tree-policy used to select the action to execute in each node needs to balance the use of already known good actions and the visitation of unknown states. The upper confidence bounds for
trees (UCT)~\cite{kocsis2006improved} algorithm is an extension for MCTS of the well-known UCB1~\cite{auer2002finite} multi-armed bandit algorithm. UCB1 chooses the arm (action $a$) using
\begin{flalign}
a = \argmax_{i \in \{1...K\}} \overline{X}_{i, T_i(n-1)} + C\sqrt{\frac{\log n}{T_i(n-1)}},\nonumber
\end{flalign}
where $T_i(n) = \sum^n_{t=1} \textbf{1} \{t=i\} $ is the number of
times arm $i$ is played up to time $n$. $\overline{X}_{i, T_i(n-1)}$
denotes the average reward of arm $i$ up to time $n-1$ and $C
= \sqrt{2}$ is an exploration constant. 
In UCT, each node is a separate bandit, where the arms correspond to the actions, and the payoff is the reward of the episodes starting from them.
In the backup phase, value is backed up recursively from the leaf node to the root as
\begin{flalign}
\overline{X}_n = \sum^{K}_{i=1} \Big(\frac{T_i(n)}{n}\Big) \overline{X}_{i, T_i(n)}.\nonumber
\end{flalign}
Kocsis et al. 2006 \cite{kocsis2006improved} proved that UCT converges in the limit to the optimal policy.
\subsection{Wasserstein barycenter}
\label{sec_wasserstein_Barycenter}
Let $(\mathcal{X}, d)$ be a complete separable metric (Polish) space and $x_0 \in \mathcal{X}$ be an arbitrary point. For each $q \in \mathbb{R}, q \geq 1$ we define ${P}_q(\mathcal{X})$ as the set of all probability measures $\mu$ over $(\mathcal{X}, {P})$. Let $\mu, \nu \in {P}_q(\mathcal{X})$. The $L^{q}$-Wasserstein distance between $\mu, \nu$ is defined as
\begin{align}
    W_q(\mu, \nu) = \bigg(\underset{\rho \in \Gamma(\mu, \nu)}{\text{inf}} \underset{X,Y \sim \rho}{\mathbb{E}}[d(X, Y)^q] \bigg)^{1/q},\nonumber
\end{align}
where $\Gamma(\mu, \nu)$ is the set of all probability measures on $\mathcal{X} \times \mathcal{X}$ (couplings) with marginals $\mu$ and $\nu$, and $d: \mathcal{X} \times \mathcal{X} \to \mathbb{R}^+_0$ is a convex function~\cite{wstein}. Given a set of probability measures $\{ \nu_i \}^n_{i = 1}$, belonging to the class $\mathcal{N}$, and a set of weights $\{w_i\}^n_{i = 1}$, such that $\sum^n_{i = 1} w_i = 1, w_i \geq 0$, the $L^q$-Wasserstein barycenter~\cite{agueh2011barycenters} is defined as:
\begin{align}\label{l1_wasserstein_Barycenter}
    \bar{\nu} = \underset{\nu \in \mathcal{N}}{\arg \inf} \bigg\{ \sum^n_{i = 1} w_i W_q(\nu, \nu_i)^q \bigg\}.\nonumber
\end{align}
Intuitively, the Wasserstein barycenter is the probability measure that minimizes the (weighted) Wasserstein distance of given probability measures. In the remainder of the paper, we consider the original use of $L^1$-Wasserstein barycenter equipped with $\alpha$-divergence distance for carrying out our theoretical and empirical analysis.
\vspace{-0.3cm}
\section{Wasserstein barycenter with \texorpdfstring{$\alpha$}\text{-divergence}}
We review the definition of $\alpha$\text{-divergence}~\cite{belousov2019entropic} for probability distributions and then introduce its usage as the distance measure in $L^1$-Wasserstein barycenter, instead of the commonly used $L^2$-distance based Wasserstein barycenter~\cite{metelli2019propagating}. Then, we describe how to use Wasserstein barycenters to propagate the uncertainty of value estimates, modeled as Gaussian distributions, across the tree.
\vspace{-0.3cm}
\subsection{\texorpdfstring{$\alpha$}\text{-divergence}}
The $f$-divergence~\cite{csiszar1964informationstheoretische} is a generalization of the distance between two points $X$ and $Y$ on a manifold $\mathcal{M}$ with coordinates $\xi^{(i)}_{X}, \xi^{(i)}_{Y}$ as
\begin{align}
    D_f \left( X \middle\| Y \right) = \sum_{i} \xi^{(i)}_{Y} f\left(\frac{\xi^{(i)}_{X}}{\xi^{(i)}_{Y}}\right),\nonumber
\end{align}
where $f$ is a convex function on $(0, \infty)$, with $f(1) = 0$.
For example, the KL-divergence corresponds to $f_{KL} = x\log x - (x-1)$.
The $\alpha-$divergence is a subclass of $f$-divergence generated by $\alpha-$function with $\alpha \in \mathbb{R}$. $\alpha-$function, defined as
\begin{align}
    f_{\alpha}(x) = \frac{(x^{\alpha} - 1) - \alpha(x-1)}{\alpha (\alpha -1)}.\nonumber
\end{align} 
Thus, we can define the $\alpha-$divergence between two points $X$ and $Y$ on a finite set $\mathcal{M}$ as
\begin{align}
    D_{f_\alpha} \left( X \middle\| Y \right) = \sum_{i} \xi^{(i)}_{Y} f_\alpha \left(\frac{\xi^{(i)}_{X}}{\xi^{(i)}_{Y}}\right),\nonumber
\end{align}
and obtain the $L^1$-Wasserstein barycenter
\begin{align}
    \bar{\nu} = \underset{\nu \in \mathcal{N}}{\arg \inf} \bigg\{ \sum^n_{i = 1} w_i W_1(\nu, \nu_i) \bigg\},\nonumber
\end{align}
with 
\begin{align}
    W_1(\nu, \nu_i) = \underset{\rho \in \Gamma(\nu, \nu_i)}{\text{inf}} \underset{X,Y \sim \rho}{\mathbb{E}}[D_{f_\alpha}(X, Y)] .\nonumber
\end{align}
In the following, we use this combination of $L^1$-Wasserstein barycenter and $\alpha$-divergence to define the value function as posterior distributions of the action-value functions modeled as Gaussians.
\vspace{-0.3cm}
\subsection{V-posterior}\label{sec_v_posterior}
It is natural to define a value node as the V-posterior computed with $L^1$-Wasserstein barycenters of the children nodes Q-posteriors, following a procedure inspired by Metelli et al. 2019~\cite{metelli2019propagating} and tailored to MCTS.
\begin{definition}[V-posterior]
Given a policy $\bar{\pi}$ and a state $s \in \mathcal{S}$, we define the V-posterior $\mathcal{V}(s)$ induced by Q-posteriors $\mathcal{Q}(s,a)$ with $a \in \mathcal{A}$ as the $L^1$-Wassertein barycenter of the $\mathcal{Q}$(s, a):
\begin{align}
    \mathcal{V}(s) \in \underset{\mathcal{V}}{\arg \inf} \bigg\{ \mathbb{E}_{a \sim \bar{\pi}(.|s)}\big[W_1(\mathcal{V}, \mathcal{Q}(s,a))\big] \bigg\}.\nonumber
\end{align}
\end{definition}
In this work, we model each node in the tree as a Gaussian distribution. We define $p = 1 - \alpha$ and derive the following.
\begin{manualproposition} {1} \label{prop_1}
Consider the V-posterior value function $\mathcal{V}(s)$ as a Gaussian: $\mathcal{N}(\overline{m}(s), \overline{\sigma}^2(s))$. Define each $\mathcal{Q}(s,a)$ as the action-value function child node of $\mathcal{V}(s)$. Each $\mathcal{Q}(s,a)$ is assumed as a Gaussian distributions $\mathcal{Q}(s,a): \mathcal{N}(m(s,a), \sigma(s,a)^2)$.
If the value function $\mathcal{V}(s)$ is defined as the Wasserstein barycenter of the action-value function $\mathcal{Q}(s,a)$, given the policy $\bar{\pi}$, we have
\begin{flalign}
    \overline{m}(s) = (\mathbb{E}_{a \sim \bar{\pi}} [m(s,a)^p])^{\frac{1}{p}} \nonumber. \\
    \overline{\sigma}(s) = (\mathbb{E}_{a \sim \bar{\pi}} [\sigma(s,a)^p])^{\frac{1}{p}}.\nonumber
\end{flalign}
\end{manualproposition}
Proposition \ref{prop_1} shows the closed form solutions of the mean and the standard deviation of the Gaussian value function $\mathcal{V}(s)$ considering it as the $L^1$-Wasserstein barycenter Q-posteriors. 
In detail, the mean and the standard deviation value of $\mathcal{V}(s)$ are the power mean of all mean and standard deviation values, respectively, of all the \textbf{$\mathcal{Q}(s,a)$} function, considering the finite set of actions. When $p=1$, we derive the expected form solutions.

We point out that our approach is not restricted to the Gaussian distribution model. We get the following result by considering each tree node as a particle model.
\begin{manualproposition} {2} \label{prop_2}
Consider the V-posterior value function $\mathcal{V}(s)$ as an equally weighted Particle model: $\overline{x_i}(s): i \in [1,M]$. $M$ is an integer and $M\geq 1$. Assume each action-value function $\mathcal{Q}(s,a)$ has $M$ particles $x_i(s,a), i \in [1, M]$.
If the value function $\mathcal{V}(s)$ is defined as the Wasserstein barycenter of the action-value function $\mathcal{Q}(s,a)$, given the policy $\bar{\pi}$, each particle $\overline{x_i}(s), i\in [1,M]$ can be estimated as
\begin{align}
    \overline{x_i}(s) &= (\E_{a \sim \bar{\pi}} [x_i(s,a)^{p}])^{1/p},\nonumber
\end{align}
\end{manualproposition}
Proposition \ref{prop_2} shows that each particle of the V-posterior value function $\mathcal{V}(s)$ can be derived as the power mean of the respective particles of all the \textbf{$\mathcal{Q}(s,a)$} function. If $p=1$, we again get the closed-form solutions as the expectation of the respective particles of all the \textbf{$\mathcal{Q}(s,a)$} functions. The results in Proposition~\ref{prop_1}, and Proposition~\ref{prop_2}  can be considered as the generalized result of Proposition A.3 in Metelli et al. 2019~\cite{metelli2019propagating}.
In the next section, we present our Wasserstein Monte-Carlo tree search (W-MCTS) algorithm assuming each tree node is a Gaussian distribution.
\vspace{-0.3cm}
\section{Wasserstein Monte-Carlo tree search}
We devise our Wasserstein Monte-Carlo tree search (W-MCTS) algorithm building upon our formulation of V-posteriors modeled as Wasserstein barycenters of action-value distributions. Since we are modeling the distributions at each node as Gaussians, we require backup operators for mean and variance. Furthermore, we propose two policy sampling strategies for action selection and tree expansion based on optimistic selection and Thompson sampling.
\vspace{-0.3cm}
\subsection{Backup operator}
Consider each V-node and Q-node in the tree as a Gaussian distribution with mean and standard deviation $V_{\textrm{m}}(s), V_{\textrm{std}}(s)$, and $Q_{\textrm{m}}(s,a), Q_{\textrm{std}}(s,a)$ respectively. 
According to Proposition~\ref{prop_1}, we have 
\begin{flalign}
    &V_{\textrm{m}}(s) = \E_{a \sim \bar{\pi}} [Q_{\textrm{m}}(s,a)], \text{and } V_{\textrm{std}}(s) = \E_{a \sim \bar{\pi}} [Q_{\textrm{std}}(s,a)].\nonumber
\end{flalign}
In our W-MCTS algorithm, we consider the expectation regarding the given policy $\bar{\pi}$ as the visitation count ratio 
and propose the following mean and standard deviation value backup operator of the V-node
\begin{flalign}    
    &\overline{V}_{\textrm{m}}(s,N(s)) \gets \left(\sum_{a} \frac{n(s,a)}{N(s)} \overline{Q}_{\textrm{m}}(s,a,n(s,a))^p\right)^{\frac{1}{p}} .\nonumber\\
    &\overline{V}_{\textrm{std}}(s,N(s)) \gets \left(\sum_{a} \frac{n(s,a)}{N(s)} \overline{Q}_{\textrm{std}}(s,a,n(s,a))^p\right)^{\frac{1}{p}}.\nonumber
\end{flalign}
where $\overline{V}_{\textrm{m}}(s,N(s))$ and $\overline{Q}_{\textrm{m}}(s,a,n(s,a))$ denote the empirical mean value for V- and Q-nodes after $N(s)$, and $n(s,a)$ visitations. $\overline{V}_{\textrm{std}}(s,N(s))$ and $\overline{Q}_{\textrm{std}}(s,a,n(s,a))$ denote the empirical estimate standard deviation for V- and Q-nodes after $N(s)$, and $n(s,a)$ visitations. We note that the visitation count ratios are also used in UCT~\cite{kocsis2006improved}, and Power-UCT~\cite{dam2019generalized}. In addition, AlphaGo~\cite{silver2016mastering} and AlphaZero~\cite{silver2017amastering, silver2017bmastering} also use visitation ratio as the policy selection at the root node after building the tree.

For the backup operators of $Q_{\textrm{m}}$ and $Q_{\textrm{std}}$, we use the Bellman operator 
\begin{flalign}
    \mathcal{Q}(s,a) = \E_{\tau(s,a)} [r(s,a)] + \gamma \E_{\tau(s,a)} [\mathcal{V}(s')],\nonumber
\end{flalign}
where $s' \sim \tau(s,a)$ is the probability transition of taking action $a$ at state $s$ to derive that 
\begin{flalign}
    &Q_{\textrm{m}}(s, a) = \E_{\tau(s,a)} [r(s,a)] + \gamma \E_{\tau(s,a)}[V_{\textrm{m}}(s')],\nonumber\\
    &Q_{\textrm{std}}(s, a) = \gamma V_{\textrm{std}}(s'),\nonumber
\end{flalign}
and employ the visitation ratio as the policy selection, we get the backup operators
\begin{flalign}    
    &\overline{Q}_{\textrm{m}}(s, a,n(s, a)) \gets \frac{\sum r(s,a) + \gamma \sum_{s'} N(s') \overline{V}_{\textrm{m}}(s',N(s'))}{n(s, a)} \nonumber. \\
    &\overline{Q}_{\textrm{std}}(s,a,n(s, a)) \gets \frac{\gamma \sum_{s'} N(s') \overline{V}_{\textrm{std}}(s',n(s, a))}{n(s, a)}. \nonumber
\end{flalign}
 Note that the mean backup operators in W-MCTS are the same as Power-UCT~\cite{dam2019generalized} using power mean with a constant $p \geq 1$. However, by considering each node as a Gaussian distribution, we further provide the standard deviation update for each tree node. We further point out that, differing from distributional RL approaches, the variances of the Q-node and V-node in W-MCTS decrease to zero when increasing the number of samples and reduce to deterministic value functions.
Next, we will present the two action selection methods employed in W-MCTS.
\vspace{-0.2cm}
\subsection{Action selection}
\subsubsection{Optimistic selection}
The first sampling method relies on the optimism in the face of uncertainty (OFU) principle, used in UCT. In UCT, an action is selected to maximize the statistical upper confidence limits of lower level $Q$-nodes as 
\begin{flalign}
a = \argmax_{a_i, i \in \{1...K\}} m(s,a_i) + C\sqrt{\frac{\log N(s)}{n_i(s,a_i)}},\nonumber
\end{flalign}
 with $i=1,2...,K$ is the action index, $N(s)$ is the visitation count of the current $V$ node at state $s$, and $n_i(s,a_i)$ is the visitation count of the $Q$ node at state $s$ action $a_i$, and $C$ is an exploration constant.
We observe that we can replace the exploration term $1/\sqrt{n_i(s,a_i)}$ by the standard deviation of the $Q(s,a_i)$ node and derive Wasserstein Monte-Carlo tree search using optimistic selection (\textit{W-MCTS-OS}):
\begin{flalign}
a = \argmax_{a_i, i \in \{1...K\}} m(s,a_i) + C\sigma_i(s,a_i)\sqrt{\log N(s)}.\nonumber
\end{flalign}
This comes from the central limit theorem as $\sigma_i(s,a_i)^2 \sim 1/n_i(s,a_i)$.
\vspace{-0.2cm}
\subsubsection{Thompson sampling}
The second method relies on Thompson sampling, in which an action is selected based on the posteriors of action-value distributions as:
\begin{flalign}
a = \argmax_{a_i, i \in \{1...K\}} \{ \theta_i \sim \mathcal{N}(m(s,a_i), \sigma^2(s,a_i)) \}.\nonumber
\end{flalign}
Here, $\mathcal{N}(m(s,a_i), \sigma^2(s,a_i))$ is a Gaussian distribution of following $Q$-nodes with mean value $m(s,a_i)$ and standard deviation value $\sigma(s.a_i)$, for each action $a_i$ at state $s$.
We denote this method as Wasserstein Monte-Carlo tree search Thompson Sampling (\textit{W-MCTS-TS}).

\section{Theoretical Analysis}
\subsection{Wasserstein non-stationary multi-armed bandit}
We consider a class of non-stationary multi-armed bandit (MAB) problems. Let us consider $K \geq 1$ arms or actions of interest with mean value $\mu_k, k \in [K]$. Let $X_{k,t}$ denote the random reward obtained by playing arm $k \in [K]$ at the time step $t$, the reward is bounded in $[0,1]$.
$\overline{X}_{k,n} = \frac{1}{n}\sum^n_{t=1}X_{k,t}$ is the average rewards collected at arm $k$ after n times. Let $\mu_{k,n}=\E[\overline{X}_{k,n}]$. Let $*$ be the superscript for all quantities related to the optimal arm. We denote $T_k(n)$ is the number of visitations of the arm $k$. We make the following assumption:
\begin{manualassumption}{1}\label{assumpt1}
     We assume that the reward sequence, $\{ X_{k,t}: t \geq 1\}$, is a non-stationary process satisfying the assumption: 
    \begin{enumerate}
    \item (Gaussian) Each arm $k$ is a Gaussian $\mathcal{N}(\mu_k, V_k/T_k(n))$. Here $\mu_k \in [0,1], V_i > 0$.
    \item (Convergence) the expectation $\mu_{k,n}$ converge to a value $\mu_k$
    \begin{flalign}
        \mu_{k} = \lim_{n \rightarrow \infty} {\E [ \overline{X}_{kn}]}. \nonumber        
    \end{flalign}
\end{enumerate}
\end{manualassumption}
Under the Assumption~\ref{assumpt1}, we consider applying Thompson Sampling strategy as the action selection method for the non-stationary multi-armed bandit (MAB) problems describes above. Therefore, at each time step $n$, an action is selected as
\begin{flalign}
a = \argmax_{a_i, i \in \{1...K\}} \{ \theta_i \sim \mathcal{N}(\overline{X}_{k,n}, V_k/T_k(n)) \}.\nonumber
\end{flalign}
Let's define $\overline{X}_{n}(p) = \Big( \sum^{K}_{k=1} \Big(\frac{T_k(n)}{n}\Big) \overline{X}^p_{k, T_k(n)}\Big)^{1/p}$ as the power mean value backup at the root node. We show theoretical results of our method as follows. To begin, under the Assumption~\ref{assumpt1}, we derive an upper bound on the expected frequency of the number of plays of a sub-optimal arm. 
\begin{manualtheorem} {1}\label{theorem1}
Consider Thompson Sampling strategy applied to a non-stationary problem where the pay-off sequence satisfies Assumption 1. Define $V = \max_{k \in [K]}\{V_k\}$. Fix $\epsilon \geq 0$. If $k$ is the index of a sub-optimal arm, then each sub-optimal arm $k$ is played in expectation at most
\begin{flalign}
\E[T_k(n)] \leq \Theta \Bigg( 1 + \frac{V\log (n\Delta_k^2/V)}{\Delta_k^2} \Bigg).\nonumber
\end{flalign}
\end{manualtheorem}
The proof of Theorem~\ref{theorem1} closely follows Theorem 4.2(\cite{jin2022finite}).
Next, we demonstrate the concentration of the expectation of the power mean backup operator $(\E\big[ \overline{X}_n(p) \big])$ towards the optimal mean value of $K$ arms ($\mu^*$), as outlined in Theorem~\ref{theorem2}.
\begin{manualtheorem}{2 (Convergence)}\label{theorem2}
Under Assumption~\ref{assumpt1}, the following holds
\begin{flalign}
&\big| \E\big[ \overline{X}_n(p) \big]  - \mu^{*} \big| \leq |\delta^*_n|\nonumber\\ &+ \Theta \Big(\frac{(K-1)(1+V\log(n\triangle^2/V))}{\triangle^2n} \Big)^{\frac{1}{p}}. \nonumber
\end{flalign}
\end{manualtheorem}
Additionally, we establish the concentration properties of the power mean backup operator $\overline{X}_n(p)$ towards the mean value of the optimal arm $\mu^*$, as shown in Theorem~\ref{theorem3}.
\begin{manualtheorem}{3 (Concentration)}\label{theorem3}
    $\exists$ constant $C_0>0, \alpha > 0, \beta > 0$ that for any $\epsilon > 0, \exists N^{'}_p>0$, that for any $n \geq N^{'}_p$ we can derive
    \begin{flalign}
        &\Pr \bigg(\overline{X}_n(p) - \mu^{*} \geq \epsilon \bigg ) \leq C_0n^{-\alpha}\epsilon^{-\beta},\nonumber\\
        &\Pr \bigg(\overline{X}_n(p) - \mu^{*} \leq -\epsilon \bigg ) \leq C_0n^{-\alpha}\epsilon^{-\beta}.\nonumber
    \end{flalign}
\end{manualtheorem}
Based upon the results of using Thompson sampling as action selection and power mean as the value backup operator on the described non-stationary multi-armed bandit problem, we derive theoretical results for W-MCTS-TS in an MCTS tree.
\vspace{-.2cm}
\subsection{Convergence in Monte-Carlo Tree Search}
We first show that the probability of not choosing an optimal action at the root node decays polynomially to zero as shown in Theorem~\ref{theorem4}.
\begin{manualtheorem}{4 (Convergence of Failure Probability)}\label{theorem4}
At the root node, let denote $a_k$ be the action returned by W-MCTS-TS at timestep n at the root node, $a_{k^*}$ is the optimal action. Then for $\epsilon>0$, $\exists C> 0, \alpha > 0, N_p > 0$ as constants that for all $n \geq N_p$, we have
\begin{flalign}
        &\Pr \bigg( a_k \neq a_{k^*} \bigg ) \leq  C n^{-\alpha}.\nonumber
\end{flalign}
\end{manualtheorem}
Next, we show the polynomial convergence of the expected estimated mean value function at the root node in Theorem~\ref{theorem5}.
\begin{manualtheorem} {5 (Convergence of Expected Payoff)}\label{theorem5}
We have at the root node $s^{(0)}$, $\exists N_0>0, \triangle = \max_{k\in[|K|]}\{\triangle_k\}$ so that the expected payoff satisfies
\begin{flalign}
&\big| \E [\overline{V}_m(s^{(0)},n)] - Q_m(s^{(0)},{a_{k^*}}) \big|\nonumber\\ &\leq \Theta \Bigg( \frac{2(|K|-1)(1 + \frac{V\log(n\triangle^2/V)}{\triangle^2})}{n} \Bigg). \nonumber
\end{flalign}
\end{manualtheorem}
Proof for all Theorems can be found in Appendix.
Our approach, W-MCTS-TS, guarantees a polynomial convergence rate to the original optimal value function at the root node. \citet{xiao2019maximum} propose MENTS which benefits from the exponential convergence to the regularized value function due to the maximum entropy regularization. However, MENTS suffers from the regularized value function's error, which creates bias and can result in incorrect action selection, particularly in high branching factor problems. Therefore, our approach provides a more accurate and reliable solution as it directly converges to the original optimal value function.

\noindent \textbf{Remark}: In the field of bandit problems, the theoretical analysis of Thompson Sampling in this literature is not yet fully understood (\cite{agrawal2012analysis}, \cite{kaufmann2012thompson}, \cite{korda2013thompson}, \cite{jin2022finite}). On the other hand, in MCTS, our paper makes a significant contribution by providing a clear convergence rate for W-MCTS with the Thompson Sampling strategy. Although the convergence of the Thompson Sampling method has been demonstrated in previous studies (such as the DNG~\cite{bai2013bayesian}, D2NG~\cite{bai2014thompson} methods), they did not offer a specific convergence rate. Therefore, we believe that our paper is the first to provide a convergence rate for MCTS with the Thompson Sampling strategy, which is an important step forward in this field of research.
\vspace{-0.3cm}
\subsection{Fully observable highly-stochastic problems}
\begin{figure*}[t]
\centering
\includegraphics[scale=.30]{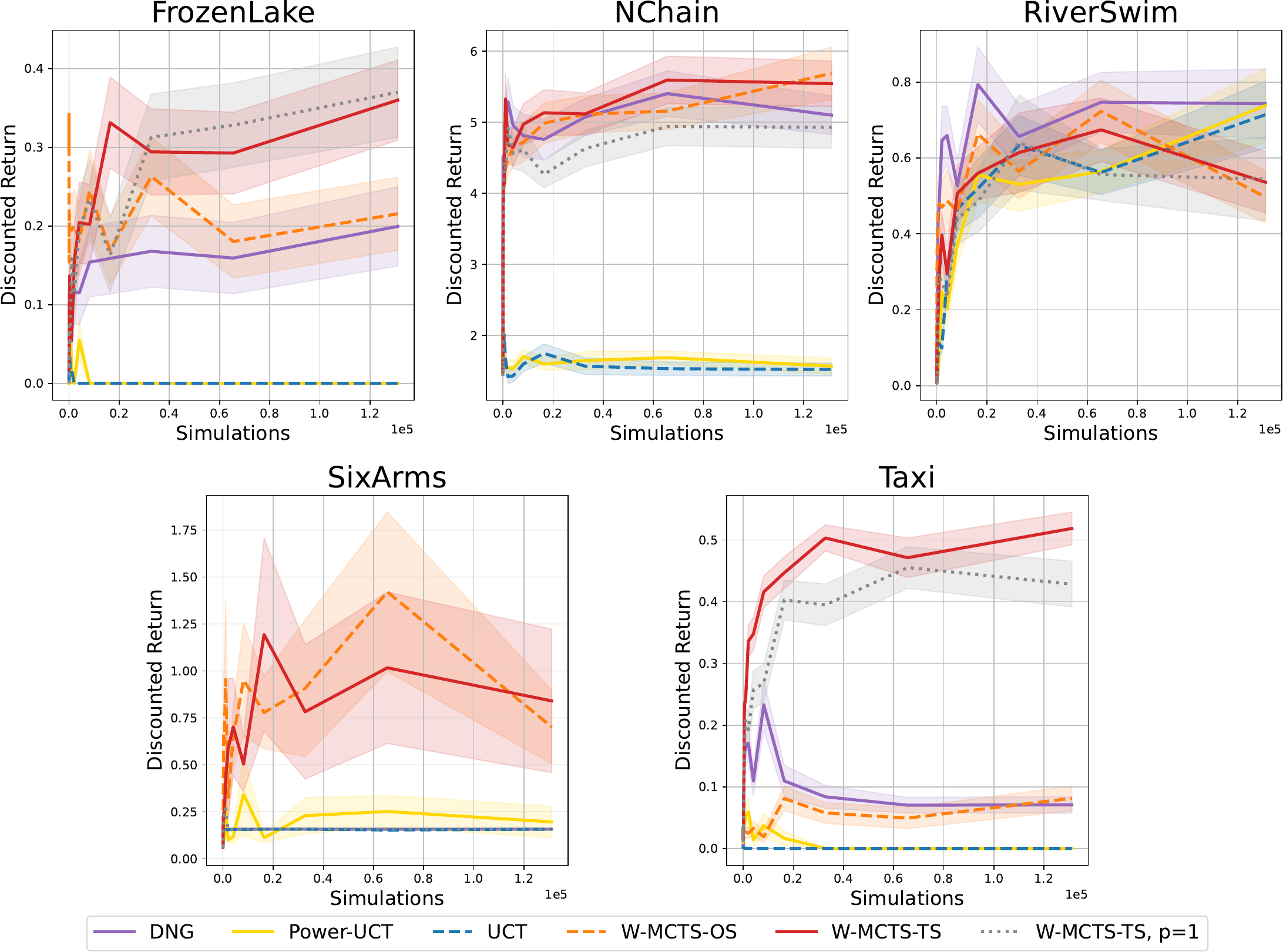}
\caption{
Performance of W-MCTS compared to DNG, Power-UCT and UCT in five different MDP environments.
The mean of total discounted reward over $50$ evaluation runs is shown by thick lines while the shaded area shows standard error.}
\label{mdp_fig_all}
\end{figure*}
We compare our methods with UCT, Power-UCT and DNG in five stochastic environments \textit{FrozenLake}, \textit{NChain}, \textit{RiverSwim, SixArm}, and \textit{Taxi} in order to understand the benefits of W-MCTS compared to baselines in highly-stochastic MDPs.
\vspace{-0.2cm}
\subsubsection{FrozenLake}
First, we consider the well-known \textit{FrozenLake} problem implemented in OpenAI Gym~\cite{brockman2016openai}. In this problem, the agent starts in the upper left corner and tries to find a way to reach a target position in the lower right corner of an 4x4 ice grids. Some holes are placed in the grid, and the agent must avoid falling or stepping on unstable spots. The task contains a high degree of stochasticity since the agent runs only one-third of the time in the intended direction and the rest of the time in one of the two tangential directions. Reaching the target position yields a +1 reward, while all other outcomes (reaching the time limit or falling into the water) yield a zero reward.

As shown in Figure~\ref{mdp_fig_all}, W-MCTS-TS outperforms DNG (uses Thompson sampling), UCT, Power-UCT, and W-MCTS with optimistic selection. W-MCTS with $p=1$ perform as well as W-MCTS-TS.

\begin{figure*}[t]
\centering
\includegraphics[scale=.33]{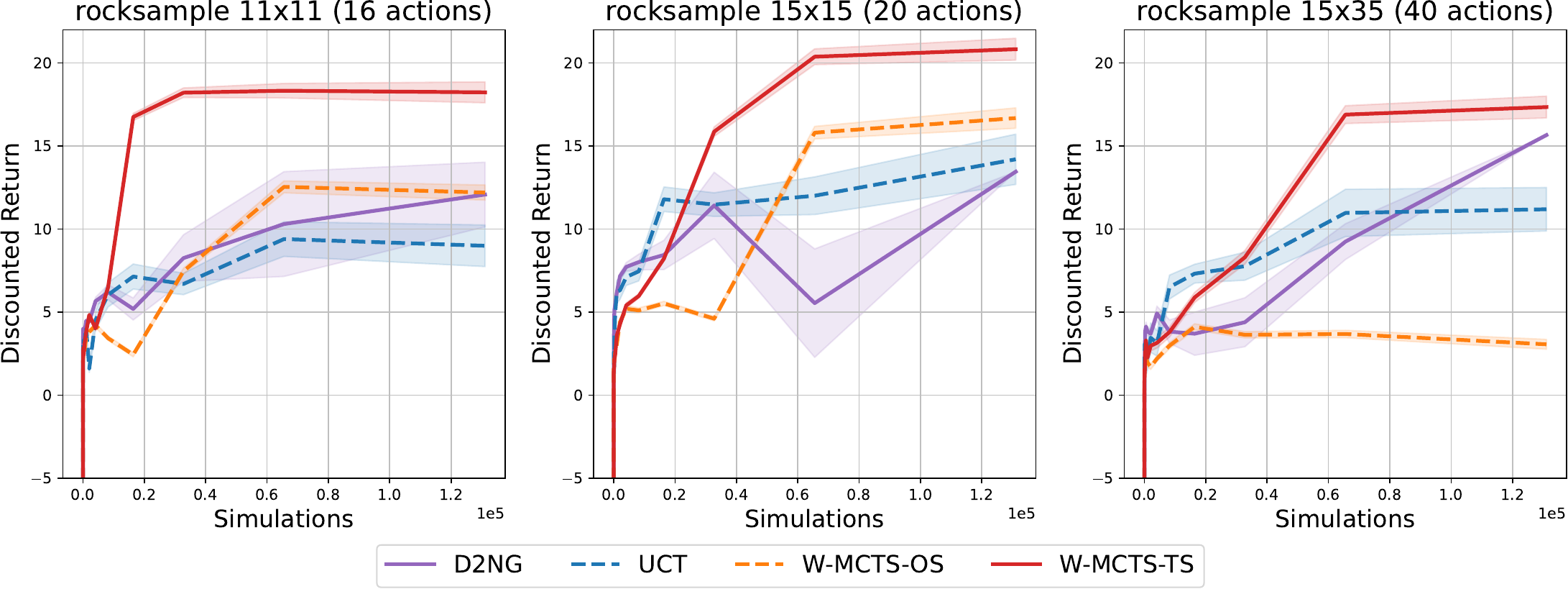}
\caption{
Performance of W-MCTS compared to D2NG in \textit{rocksample}.
The mean of total discounted reward over $1000$ (except for UCT with $100$) evaluation runs is shown by thick lines while the shaded area shows standard error.}
\label{rocksample_fig_all}
\end{figure*}
\vspace{-0.3cm}
\subsubsection{NChain}
Second, we consider the \textit{NChain} environment \cite{chainenv} of length $5$. In this environment, an agent starts at the beginning of a chain of states and can either move along the chain to collect rewards. There are only two available actions: moving backward (which will lead the agent back to the initial state), and moving forward. There is a probability of $20\%$ that the inverse action is performed instead. We use a random rollout policy to evaluate newly added value nodes. This environment acquires a high exploration strategy.

In NChain, The results illustrated in Figure~\ref{mdp_fig_all} clearly show that both Wasserstein sampling methods outperform UCT and Power-UCT. In detail, Thompson sampling shows a faster convergence rate, while, Optimistic selection has an overall higher peak. On the other hand, Power-UCT seems to struggle with this environment.
 \vspace{-0.2cm}
\subsubsection{RiverSwim}
Third, we study RiverSwim \cite{riverandsix} which consists of 5 states. It is very similar to the NChain, with the difference that moving back will only move the agent back to one state, and it is also possible for the agent to stay in the current state after trying to move in the direction of the highest reward. Additionally, the selected action is always performed, but the one in the direction of the highest reward underlies stochasticity regarding the state transitions. We used a random rollout policy to initiate the value of newly added nodes. The difficulty here is that it takes a lot more time to reach the final state when compared to the \textit{NChain} environment, so the agent has to be able to plan sufficiently well in advance. However, once it has reached the final state, it is quite unlikely that it will ever go back or fail to collect the higher reward.

On RiverSwim (see Figure~\ref {mdp_fig_all}), W-MCTS using Optimistic selection converges the fastest and obtains the best results, whereas Power-UCT, while converging the slowest, also reaches similar peaks.
\vspace{-0.3cm}
\subsubsection{SixArms}
Fourth, we consider \textit{SixArms} \cite{riverandsix}, which is composed of $7$ states, including the initial state. In this environment, an agent can move to one of $6$ states (simulating $6$ arms bandit). However, this movement underlies stochasticity, as each of the $6$ actions leading to their respective state has a different probability of success. Once the agent moves into one of those states, it can subsequently collect a reward that scales inversely with the probability of getting to that state. This environment is highly stochastic and requires lots of exploration. On \textit{SixArms}, W-MCTS shows its potential on environments with very high variance again as it is the only algorithm that obtains good results, as seen in Figure~\ref{mdp_fig_all}.
\vspace{-.2cm}
\subsubsection{Taxi}
 The fifth environment is \textit{Taxi} \cite{chainenv}. In this environment, the agent starts in the top left of a 7x6 grid and has to reach the top right. However, some tiles are walls on which it is impossible to move, and just reaching the end will not yield any reward. For that, the agent must first collect three passengers spread over the grid. A different reward is obtained depending on how many passengers the agent can collect and get to the target position. To simulate the random traffic conditions, the agent can also slip like in FrozenLake. In this case, this means that in $10\%$ of the time, the agent will not perform the intended action but instead one perpendicular to it at random. This highly stochastic environment requires lots of exploration.

 In this environment, we modified the random rollout policy to put heuristic knowledge that the rollout policy only considers actions that, barring any slipping, would move the agent in some direction.
The \textit{Taxi} environment turned out to be one of the hardest to solve successfully, and only W-MCTS, when combined with Thompson sampling, can handle the large state space well enough (see Figure~\ref{mdp_fig_all}) as it is the only algorithm that managed to pick up all three passengers. Both optimistic selection and Power-UCT struggle to search over the grid and perform poorly compared to W-MCTS with Thompson sampling.
\vspace{-0.3cm}
\subsection{Partially observable highly-stochastic problems}
\subsubsection{Rocksample}
In \textit{rocksample (n,k)} (\cite{smith2004heuristic}), a robot navigates in a fixed size \textit{$n \times n$} grid containing \textit{k} rocks. These rocks can be valuable and invaluable and are unknown to the robot.
The robot's goal is to collect valuable rocks (by sampling) and leave the grid to the east. The robot can select four navigation actions (North, South, East, West) and/or sample/not sample one of the $k$ rocks.
Therefore, there are $k + 5$ actions in total. This environment needs a robust exploration policy to collect long-term rewards.
In this task, we measure our method in three variants with a different number of actions: \textit{rocksample} (11,11) (16 actions), \textit{rocksample} (15,15) (20 actions), \textit{rocksample} (15,35) (40 actions). Results in Figure~\ref{rocksample_fig_all} indicate that W-MCTS with Thompson sampling outperforms both UCT and D2NG in all three settings.
\vspace{-.2cm}
\subsubsection{Pocman}
We evaluate our algorithm in another POMDP, the \textit{pocman} problem~\cite{silver2010monte}. In \textit{pocman}, an agent named PocMan navigates a maze of size (17x19). PocMan does not observe the entire map of the environment and can only observe the local neighborhoods of nearby cells. 
The task is to navigate the map and eat as many food pellets as possible while four ghosts move around the map trying to kill PocMan. The four ghosts move randomly at first, but then head for the area where a high number of food pellets is more likely to be found. PocMan receives a reward of $-1$ for each step he takes, $+10$ for eating each food pellet, $+25$ for eating a ghost, and $-100$ for dying. The \textit{pocman} problem has $4$ actions, $1024$ observations, and about $10^{56}$ states. Table~\ref{pocman_table} shows that
W-MCTS Thompson sampling (p=100) outperforms both UCT and D2NG in the majority of cases with a number of samples of $4096, 32768, 65536$.
\begin{table}
\caption{Discounted total reward in \textit{pocman} for the comparison methods. Mean $\pm$ standard error are computed from $1000$ random seeds.}\smallskip
\centering
\resizebox{\columnwidth}{!}{
\smallskip
\scalebox{0.7}{
\begin{tabular}{|l|c|c|c|c|}\hline
 & $1024$ & $4096$ & $32768$ & $65536$ \\\cline{1-5}
 $\text{W-MCTS-OS}, p=1$ & $50.9 \pm 0.6$ & $51.0 \pm 0.62$ & $52.2 \pm 0.79$ & $54.6 \pm 1.08$\\\cline{1-5}
$\text{W-MCTS-TS}, p=100$ & $67.38 \pm 0.53$ & $\mathbf{75.64 \pm 0.51}$ & $\mathbf{77.68 \pm 0.77}$ & $\mathbf{77.70 \pm 1.22}$\\\cline{1-5}
$\text{D2NG}$ & $\mathbf{71.55 \pm 0.57}$ & $75.39 \pm 1.47$ & $76.90 \pm 6.40$ & $72.2 \pm 0.0$\\\cline{1-5}
$\text{UCT}$ & $23.4 \pm 0.99$ & $23.6 \pm 1.09$ & $24.90 \pm 3.40$ & $28.5 \pm 3.8$\\\cline{1-5}
\end{tabular}}}
\label{pocman_table}
\end{table}
\vspace{-.1cm}
\section{Conclusion}
This work introduced a novel probabilistic approach for Monte-Carlo tree search(MCTS) designed for highly stochastic problems, which propagates the uncertainty of the estimate of value nodes across the tree. Our approach models value nodes as Gaussian distributions and adopts a novel backup operator that computes value nodes as Wasserstein barycenters of children action-value nodes. For this reason, we call our method Wasserstein Monte-Carlo Tree Search (W-MCTS). Additionally, by studying $L^1$-Wasserstein barycenters in combination with $\alpha$-divergence measure, our approach enjoys the advantageous properties of generalized mean backup~\cite{dam2019generalized} to counteract the inaccuracy in the value estimate caused by high stochasticity. We considered two action selection strategies, namely Thompson sampling and optimistic selection. We theoretically prove the polynomial convergence of W-MCTS with Thompson sampling to choose the optimal action at the root node, and empirically validated the approach in several stochastic MDPs and POMDPs.

\bibliography{main}
\bibliographystyle{icml2023}
\onecolumn
\appendix
\subsection{Outline}
\begin{itemize}
    \item Notations will be described in Section A.
    \item Hyperparameters are provided in Section B.
    \item Derivation of Wasserstein barycenter with Gaussian and particle filter distributions will be described in Section C.
    \item Full proof for the convergence of Wasserstein Non-stationary multi-armed bandit will be provided in Section D.
    \item Full proof for the convergence of Wasserstein Monte-Carlo tree search will be provided in Section E.
\end{itemize}
\vspace{-.5cm}
\subsection{A. Notations}
\vspace{-.4cm}
\begin{table}[!ht]
\caption{List of all notations of Wasserstein barycenter with Gaussian and particle filter distributions.}
\centering
\renewcommand*{\arraystretch}{2}
\begin{tabular}{ccc} 
    \toprule
    \textbf{Notation} &  \textbf{Type} & \textbf{Description}\\ \hline
    \midrule
 $\mathcal{N}(m, \delta^2)$ & $\mathbb{R}$ & Gaussian distribution with mean $m$, standard deviation $\delta$\\
 \hline
$(\mathcal{X}, d)$ &  & complete separable metric (Polish) space\\
 \hline
 $W_q(\mu, \nu)$ &  & $L^{q}$-Wasserstein distance between $\mu, \nu$\\
 \hline
 $W_1(\mu, \nu)$ &  & $L^{1}$-Wasserstein distance between $\mu, \nu$\\
 \hline
 $F^{-1}_{p(x)}(t)$ & & quantile function of a distribution $p(x)$\\
 \hline
 $\Gamma(\mu, \nu)$ & $\mathcal{X} \times \mathcal{Y}$ & set of measures on $\mathcal{X} \times \mathcal{Y}$ with marginals $\mu, \nu$ \\
 \hline
 $d(X,Y)$ & $\mathbb{R}$ & distance between $X$ and $Y$\\
 \hline
 $D_{f_\alpha}(X||Y)$ & $\mathbb{R}$ & $\alpha$-divergence distance between $X$ and $Y$\\
 \hline
 $\text{erf}^{-1}(t)$ & & $\text{ the inverse of the function }\frac{2}{\sqrt{\pi}}\int^{t}_0 \exp\{ -x^2 \} dx$\\
 \hline
\bottomrule
\end{tabular}\label{list_notations_Wasserstein}
\end{table}
\vspace{-.5cm}
\begin{table}[!ht]
\caption{List of all notations of Wasserstein Non-stationary multi-armed bandits.}
\centering
\renewcommand*{\arraystretch}{2}
\begin{tabular}{ccc} 
    \toprule
    \textbf{Notation} &  \textbf{Type} & \textbf{Description}\\ \hline
    \midrule
 $K$ & $\mathbb{N}$ & number of arms/actions\\
 \hline
 $\mu_k$ & $\mathbb{R}$ & mean value of arm $k$\\
 \hline
 $\mu^*$ & $\mathbb{R}$ & optimal mean value\\
 \hline
 $\triangle_k$ & $\mathbb{R}$ & $\triangle_k = \mu^* - \mu_k$\\
 \hline
 $\triangle$ & $\mathbb{R}$ & $\triangle = \max_{k\in[K]}\{\triangle_k\}$\\
 \hline
 $\overline{\mu}^{*}_s$ & $\mathbb{R}$ &  average reward of the optimal arm after $s$ visitations\\
 \hline
 $F^*_{s}$ & $\mathbb{R}$ &  CDF of Gaussian with mean $\overline{\mu}^{*}_s$\\
 \hline
 $T_k(n)$ & $\mathbb{N}$ & number of visitations of arm $k$ at timesteps $n$ \\
 \hline
 $\overline{X}_n(p)$ & $\mathbb{R}$ & power mean backup operator with power $p$\\
 \hline
  $\overline{X}_{k, T_k(n)}$ & $\mathbb{R}$ & average rewards of arm $k$ after $T_k(n)$ visits\\
 \hline
\bottomrule
\end{tabular}\label{list_notations_w_bandits}
\end{table}

\begin{table}[!ht]
\caption{List of all notations of Wasserstein Monte-Carlo Tree Search.}
\centering
\renewcommand*{\arraystretch}{2}
\begin{tabular}{ccc} 
    \toprule
    \textbf{Notation} &  \textbf{Type} & \textbf{Description}\\ \hline
    \midrule
 $\text{KL}$ &  & \text{KL} divergence\\
 \hline
 ${V}_m^{(0)}(s^{(0)})$ & $\mathbb{R}$ & optimal mean of V value at root state $s^{(0)}$\\
 \hline
 ${Q}_m^{(0)}(s^{(0)}, a_k)$ & $\mathbb{R}$ & \text{ mean of Q value function at state $s^{(0)}$, action $a_k$}\\
 \hline
 $\overline{V}_m^{(0)}(s^{(0)},n)$ & $\mathbb{R}$ & empirical estimated mean of V value at root state $s^{(0)}$ after n visitations\\
 \hline
 $\overline{Q}_m^{(0)}(s^{(0)},a_k,n)$ & $\mathbb{R}$ & empirical estimated mean of Q value at root at state $s^{(0)}$, action $a_k$ after n visitations\\
 \hline
 ${V}_m^{(1)}(s^{(1)})$ & $\mathbb{R}$ & optimal mean of V value at depth $(1)$ at root state $s^{(1)}$\\
 \hline
 ${Q}_m^{(1)}(s^{(1)}, a_k)$ & $\mathbb{R}$ & \text{ mean of Q value function at depth $(1)$ at state $s^{(1)}$, action $a_k$}\\
 \hline
 $\overline{V}_m^{(1)}(s^{(1)},n)$ & $\mathbb{R}$ & empirical estimated mean of V value at depth $(1)$ at state $s^{(1)}$ after n visitations\\
 \hline
 $\overline{Q}_m^{(1)}(s^{(1)},a_k,n)$ & $\mathbb{R}$ & empirical estimated mean of Q value at depth $(1)$ at state $s^{(1)}$, action $a_k$ after n visitations\\
 \hline
 $T^{(0)}_{s^{(0)},a_k}(n)$ & $\mathcal{N}$ & number of plays of action $a_k$ at state $s^{(0)}$ at timestep $n$\\
 \hline
\bottomrule
\end{tabular}\label{list_notations_w_mcts}
\end{table}
\subsection{B. Hyperparameters}
To compare the performance of W-MCTS to other state-of-the-art planning algorithms, we run several experiments on standard MDP as well as POMDP environments. For comparison, we consider UCT~\cite{kocsis2006improved}, Power-UCT~\cite{dam2019generalized}, DNG~\cite{bai2013bayesian} and D2NG~\cite{bai2014thompson}. The hyperparameters are tuned using grid-search. Except for the case of PocMan environment, we scale the rewards into the range [0, 1]. We use the discount factor $\gamma = 0.95$. For DNG, D2NG, we set hyperparameters as recommended in the paper, and from author source code. We set exploration constant for UCT, Power-UCT to $\sqrt{2}$. We set initial standard deviation value to $std=30.$ In all Rocksample and Pocman environments, we set the heuristic for rollouts as $treeknowledge =0, rolloutknowledge =1$.
For all environments we increase the value of $p$ and choose the best power mean $p$ value for Power-UCT, and W-MCTS. Details can be found in the table below. 
\begin{table}[!ht]
\caption{List of all hyperparameters.}
\centering
\begin{tabular}{ccc} 
    \toprule
    \textbf{Environments} &  \textbf{p Value Search} & \textbf{Best p Value}\\ \hline
    \midrule
 FrozenLake & $p=1,2,4,10,100$ & W-MCTS-OS(p=100),W-MCTS-TS(p=100),Power-UCT(p=100)\\
 \hline
 NChain & $p=1,2,4,8,15,100$ & W-MCTS-OS(p=4),W-MCTS-TS(p=100),Power-UCT(p=8)\\
 \hline
 RiverSwim & $p=1,2,4,8,15,100$ & W-MCTS-OS(p=100),W-MCTS-TS(p=15),Power-UCT(p=15)\\
 \hline
 SixArms & $p=1,2,4,8,15,100$ &  W-MCTS-OS(p=100),W-MCTS-TS(p=100),Power-UCT(p=8)\\
 \hline
 Taxi & $p=1,2,4,8,15,100$ & W-MCTS-OS(p=15),W-MCTS-TS(p=15),Power-UCT(p=15)\\
 \hline
 Rocksample(11x11) & $p=10,50,80,100,150$ &  W-MCTS-OS(p=150),W-MCTS-TS(p=100)\\
 \hline
 Rocksample(15x15) & $p=10,50,80,100,150$ &  W-MCTS-OS(p=100),W-MCTS-TS(p=100)\\
 \hline
 Rocksample(15x35) & $p=10,80,100$ &  W-MCTS-OS(p=150),W-MCTS-TS(p=10)\\
 \hline
 Pocman & $p=1,2,4,8,10,100$ &  W-MCTS-OS(p=1),W-MCTS-TS(p=100)\\
 \hline
\bottomrule
\end{tabular}\label{hyperparams}
\end{table}
\subsection{C. Derivation of Wasserstein barycenter with Gaussian and particle filter distributions}
We revisit the definition of Wasserstein distance:
The $L^{q}$-Wasserstein distance (with $q>0$) between two distributions $\mu, \nu$ with the cost function $d(x,y): \mathcal{X} \times  \mathcal{Y} \rightarrow \R$ is defined as
\begin{align}
    W_q(\mu, \nu) = \bigg(\underset{\rho \in \Gamma(\mu, \nu)}{\text{inf}} \underset{X,Y \sim \rho}{\mathbb{E}}[d(X, Y)^q] \bigg)^{1/q}, 
\end{align}
here $\Gamma(\mu, \nu)$ is the set of measures on $\mathcal{X} \times \mathcal{Y}$ with marginals $\mu, \nu$.\\
Define $F^{-1}_{p(x)}(t)$ as the quantile function of a distribution 
\begin{flalign}
    p(x): F^{-1}_{p(x)}(t) = \inf \{x\in \R, t \leq F_{p}(x)\}.
\end{flalign}
With $d(X,Y) = |X-Y|$ as the Euclidean distance, we can derive
\begin{align}
    W^q_q(\mu, \nu) = \bigg(\int^1_0|F^{-1}_{\mu}(t) - F^{-1}_{\nu}(t)|^q dt \bigg).
\end{align}
With $d(X,Y) = D_{f_\alpha}(X||Y),$ as the $\alpha$-divergence distance (defined in section 4.1), we can derive
\begin{align}
    W^q_q(\mu, \nu) = \bigg(\int^1_0 D_{f_\alpha}(F^{-1}_{\mu}(t)|| F^{-1}_{\nu}(t))^q dt \bigg). \label{wasser-alpha-dist}
\end{align}
\subsubsection{\texorpdfstring{$L^1$}\text{-Wasserstein} barycenter with \texorpdfstring{$\alpha$}\text{-divergence distance}}
We have 
\begin{align}
    W_1(\mu, \nu) = \underset{\rho \in \Gamma(\mu, \nu)}{\text{inf}} \underset{X,Y \sim \rho}{\mathbb{E}}[d(X, Y)] = \underset{\rho \in \Gamma(\mu, \nu)}{\text{inf}} \underset{X,Y \sim \rho}{\mathbb{E}}[D_{f_\alpha}(X, Y)] \label{w_1_alpha_divergence}.
\end{align}
We find the lower bound of $W_1(\mu, \nu)$ with $\alpha$-divergence as a measure cost function.\\
Let denote $\mathcal{N}(m, \delta^2)$ as a Gaussian distribution with mean $m$ and standard deviation $\delta$. With $\mu = \mathcal{N}(m_1,\delta_1^2), \nu = \mathcal{N}(m_2, \delta_2^2) $
We first want to show that by applying Data Processing Inequalities (Lemma 2.1~\cite{gerchinovitz2020fano}), with $h(X) = X-m_1$, and $g(X) = X-m_2$, we can derive
\begin{flalign}
&W_1 (\mu, \nu) = \underset{\rho \in \Gamma(\mu, \nu)}{\text{inf}} \underset{X,Y \sim \rho}{\mathbb{E}}[D_{f_\alpha}(X, Y)]] \geq \underset{\rho \in \Gamma(\mu, \nu)}{\text{inf}} \underset{X,Y \sim \rho}{\mathbb{E}} [ D_{f_\alpha}(X - m_1, Y - m_1)]\label{m2_m1} \nonumber \\
&= W_1 (\mathcal{N}(0,\delta_1^2), \mathcal{N}(m_2-m_1,\delta_2^2)),\\
&\text{and}\nonumber\\
&W_1 (\mu, \nu) = \underset{\rho \in \Gamma(\mu, \nu)}{\text{inf}} \underset{X,Y \sim \rho}{\mathbb{E}}[D_{f_\alpha}(X, Y)]] \geq \underset{\rho \in \Gamma(\mu, \nu)}{\text{inf}} \underset{X,Y \sim \rho}{\mathbb{E}} [ D_{f_\alpha}(X - m_2, Y - m_2)]\nonumber\\
&\geq \underset{\rho \in \Gamma(\mu, \nu)}{\text{inf}} \underset{X,Y \sim \rho}{\mathbb{E}} [ D_{f_\alpha}(m_2 - X, m_2 - Y)] (\text{ with the transform function $f(X)=-X$})\nonumber \\
&= W_1 (\mathcal{N}(m_2 - m_1,\delta_1^2), \mathcal{N}(0,\delta_2^2)). \label{m1_m2}
\end{flalign}
Now according to (\ref{wasser-alpha-dist}), the $L^1$-Wasserstein distance with $\alpha$-divergence distance is defined as 
\begin{align}
    W_1(\mu, \nu) = \bigg(\int^1_0 D_{f_\alpha}(F^{-1}_{\mu}(t)|| F^{-1}_{\nu}(t)) dt \bigg).\label{l1_w1_distance}
\end{align}
We show that the quantile function of a Gaussian distribution~\cite{Sochstatsproofs} $F = \mathcal{N}(\mu, \delta^2)$ is 
\begin{flalign}
    F^{-1}(t) = \sqrt{2}\delta \text{erf}^{-1}(2t - 1) + \mu,
\end{flalign}
where $\text{erf}^{-1}(t)\text{ is the inverse of the function }\frac{2}{\sqrt{\pi}}\int^{t}_0 \exp\{ -x^2 \} dx$.\\
Therefore, the  $L^1$-Wasserstein distance with $\alpha$-divergence distance as the cost function between two Gaussian distributions $\mu = \mathcal{N}(m_1,\delta_1^2), \nu = \mathcal{N}(m_2, \delta_2^2) $ can be measured as
\begin{align}
   W_1(\mu, \nu) = \bigg(\int^1_0 D_{f_\alpha}(\sqrt{2}\delta_1 \text{erf}^{-1}(2t - 1) + m_1|| \sqrt{2}\delta_2 \text{erf}^{-1}(2t - 1) + m_2) dt \bigg). \nonumber
\end{align}
Applying the convexity properties of $\alpha$-divergence~\cite{Amari_family_divergence}, and from (\ref{m2_m1}),(\ref{m1_m2}) we have
\begin{flalign}
    &W_1 (\mu, \nu) \geq \frac{1}{2} \bigg(\int^1_0 D_{f_\alpha}(\sqrt{2}\delta_1 \text{erf}^{-1}(2t - 1)|| \sqrt{2}\delta_2 \text{erf}^{-1}(2t - 1) + m_2-m_1) dt \nonumber\\
    &+ \int^1_0 D_{f_\alpha}(\sqrt{2}\delta_1 \text{erf}^{-1}(2t - 1) + m_2 - m_1|| \sqrt{2}\delta_2 \text{erf}^{-1}(2t - 1)) dt \bigg) \nonumber\\
    &\geq \bigg(\int^1_0 D_{f_\alpha} (\sqrt{2}\delta_1 \text{erf}^{-1}(2t - 1) + \frac{m_2-m_1}{2} || \sqrt{2}\delta_2 \text{erf}^{-1}(2t - 1) + \frac{m_2-m_1}{2}) \bigg)\nonumber\\
    &= W_1 (\mathcal{N}(\frac{m_2 - m_1}{2},\delta_1^2), \mathcal{N}(\frac{m_2 - m_1}{2},\delta_2^2)).\nonumber
\end{flalign}
Applying Data Processing Inequalities (Lemma 2.1~\cite{gerchinovitz2020fano}), with $h(X) = X-\frac{m_2-m_1}{2}$, we can derive
\begin{flalign}
    W_1 (\mu, \nu) \geq W_1 (\mathcal{N}(0,\delta_1^2), \mathcal{N}(0,\delta_2^2)) = \bigg(\int^1_0 D_{f_\alpha}(\sqrt{2}\delta_1 \text{erf}^{-1}(2t - 1)|| \sqrt{2}\delta_2 \text{erf}^{-1}(2t - 1)) dt \bigg). \nonumber
\end{flalign}
Let us consider the sequences $0 = t_0 \leq t_1 \leq ... \leq t_N=1$, there exists $\xi_i \in [t_i, t_{i+1}]$ that
\begin{align}
   W_1(\mu, \nu) &\geq \sum^{i = N}_{i=0} (t_{i+1} - t_i) D_{f_\alpha}(\sqrt{2}\delta_1 \text{erf}^{-1}(2\xi_i - 1)|| \sqrt{2}\delta_2 \text{erf}^{-1}(2\xi_i - 1))\nonumber\\
   &=\sum^{i = N}_{i=0} \Delta_i D_{f_\alpha}(\sqrt{2}\delta_1 \text{erf}^{-1}(2\xi_i - 1)|| \sqrt{2}\delta_2 \text{erf}^{-1}(2\xi_i - 1)),\nonumber
\end{align}
with  $\Delta_i = (t_{i+1} - t_i)$.
Since $D_{f_\alpha}(cP||cQ) = D_{f_\alpha}(P||Q)$ where $c$ is a constant. We can derive
\begin{align}
   W_1(\mu, \nu) \geq \sum^{i = N}_{i=0} \Delta_i D_{f_\alpha}(\delta_1 || \delta_2) = D_{f_\alpha}(\delta_1 || \delta_2). \label{w1_std}
\end{align}
We start with the first Proposition about the closed solutions of mean and variance of a Gaussian value function $\mathcal{V}(s)$ as V-posterior $L^1$-Wasserstein barycenter of all action value function distributions $\mathcal{Q}(s,a)$.
\begin{manualproposition} {1}
Consider the V-posterior value function $\mathcal{V}(s)$ as a Gaussian: $\mathcal{N}(\overline{m}(s), \overline{\delta}^2(s))$. Let's define each $\mathcal{Q}(s,a)$ as the Q function child node of $\mathcal{V}(s)$. Each $\mathcal{Q}(s,a)$ is assumed as a Gaussian distributions $\mathcal{Q}(s,a): \mathcal{N}(m(s,a), \delta(s,a)^2)$.
If the value function $\mathcal{V}(s)$ is defined as the Wasserstein barycenter of the Q function $\mathcal{Q}(s,a)$ given the policy $\bar{\pi}$, we will have:
\begin{flalign}
    \overline{m}(s) = (\mathbb{E}_{a \sim \bar{\pi}} [m(s,a)^p])^{\frac{1}{p}}\\
    \overline{\delta}(s) = (\mathbb{E}_{a \sim \bar{\pi}} [\delta(s,a)^p])^{\frac{1}{p}},
\end{flalign}
with $p = 1 - \alpha$.
\end{manualproposition}

\begin{proof}
By the definition of the V-posterior value function, we have:
\begin{flalign}
    (\overline{\mu}(s), \overline{\delta}(s)) = \argmin_{\mu,\delta} \Big \{ \E_{\bar{\pi}}[W_1(\mathcal{V}(s) || \mathcal{Q}(s,a))] \Big \}. \label{w1_v_posterior}
\end{flalign}
We first compute the standard deviation $\overline{\delta}(s)$.\\
From (\ref{w1_std}), and (\ref{w1_v_posterior}), we want to find $\overline{\delta}(s)$ that is the minimizer of 
\begin{flalign}
    \overline{\delta}(s) = \argmin_{\delta(s)} \Big \{ \E_{\bar{\pi}}[D_{f_\alpha}(\delta(s) || \delta(s,a))] \Big \}.\nonumber
\end{flalign}
 we derive
$\overline{\delta}(s)$ is the solution of
\begin{flalign}
    \frac{\nabla \E_{a \sim \bar{\pi}}[D_{f_\alpha}(\delta(s) || \delta(s,a))]}{\nabla \delta(s)} = 0.
\end{flalign}
Since
\begin{equation}
   \frac{\nabla f_{\alpha}(x)}{\nabla x}  = \frac{\alpha(x^{\alpha-1} - 1)}{\alpha(\alpha-1)} = \frac{x^{\alpha-1} - 1}{\alpha-1}.
\end{equation}
With $D_{f_\alpha}(x||y) = \sum_y y f_{\alpha}(\frac{x}{y}) $, we can have
\begin{flalign}
   \frac{\nabla D_{f_\alpha}(x||y)}{\nabla x}  = \sum_y \frac{(\frac{x}{y})^{\alpha-1} - 1}{\alpha-1}.
\end{flalign}
We can derive
\begin{equation}
\E_{a \sim \bar{\pi}}\Bigg[\frac{(\frac{\overline{\delta}(s)}{\delta(s,a)})^{\alpha-1} - 1}{(\alpha-1)} \Bigg] = 0
\Longrightarrow \E_{a \sim \bar{\pi}}\Bigg[(\frac{\overline{\delta}(s)}{\delta(s,a)})^{\alpha-1 }-1 \Bigg] = 0.
\end{equation}
Now we can define $p=1-\alpha$ 
that leads to 
\begin{flalign}
    \overline{\delta}(s) = (\mathbb{E}_{a \sim \bar{\pi}} [\delta(s,a)^p])^{\frac{1}{p}}.
\end{flalign}
To compute $\bar{\mu}(s)$. Let's revisit here again the definition of $L^1-$Wasserstein distance between two Gaussian distributions $\mu(m_1, \delta_1^2)$, $\nu(m_2, \delta_2^2)$.
\begin{flalign}
    W_1(\mu, \nu) = \inf \{ \E[D_{f_{\alpha}} (\mu || \nu)] \}.
\end{flalign}
According to Jensen's inequality\cite{PERLMAN197452} we can derive
\begin{flalign}
    \E[D_{f_{\alpha}} (\mu || \nu)] \geq  D_{f_{\alpha}} (\E[\mu] || \E[\nu]) = D_{f_{\alpha}} (m_1 || m_2).
\end{flalign}
Therefore, according to the definition of Wasserstein barycenter, the mean of a Gaussian V-posterior value function $\mathcal{V}(s)$ can be derived as
\begin{flalign}
    \overline{m}(s) = \argmin_{m(s)} \E_{a \sim \bar{\pi}} [ D_{f_{\alpha}} (m(s) || m(s,a)) ].
\end{flalign}
Following the same steps as to compute $\delta(s)$, we can get
\begin{flalign}
    \overline{m}(s) = (\mathbb{E}_{a \sim \bar{\pi}} [m(s,a)^p])^{\frac{1}{p}},
\end{flalign}
with $p = 1-\alpha$ that concludes the proof.\\
\end{proof}
Next, we consider each node as an equally weighted Particle model and derive the following proposition.
\begin{manualproposition} {2}
Let's assume the V-posterior value function $\mathcal{V}(s)$ as a equally weighted Particle model: $\overline{x_i}(s): i \in [1,M]$. $M$ is an integer and $M\geq 1$. Let's assume each Q function $\mathcal{Q}(s,a)$ has $M$ particles $x_i(s,a), i \in [1, M]$.
If the value function $\mathcal{V}(s)$ is defined as the Wasserstein barycenter of the Q function $\mathcal{Q}(s,a)$ given the policy $\bar{\pi}$, each particle ($\overline{x_i}(s), i\in [1,M]$) can be estimated as
\begin{align}
    \overline{x_i}(s) &= (\E_{a \sim \bar{\pi}} [x_i(s,a)^{p}])^{1/p},
\end{align}
with $p = 1 - \alpha$.
\end{manualproposition}
\begin{proof}
    
We can compute the quantile function of $\mu$ and $\nu$ as    
\begin{flalign}
    F_{\mu}^{-1}(t) = \sum^M_{i=1} x_i \1_{I_i}(t), F_{\nu}^{-1}(t) = \sum^M_{i=1} y_i \1_{I_i}(t).
\end{flalign}
Therefore from (\ref{l1_w1_distance}) we can get
\begin{align}
    W_1(\mu, \nu) &= \bigg(\int^1_0 D_{f_\alpha}(F^{-1}_{\mu}(t)|| F^{-1}_{\nu}(t)) dt \bigg)\\
    &=\sum^M_{i = 1}\bigg(\int_{I_i} D_{f_\alpha}(F^{-1}_{\mu}(t)|| F^{-1}_{\nu}(t)) dt \bigg)\\
    &= \sum^M_{i = 1}\bigg(\int_{I_i} D_{f_\alpha}(x_i|| y_i) dt \bigg)\\
    &= \sum^M_{i = 1}D_{f_\alpha}(x_i|| y_i) \bigg(\int_{I_i}  dt \bigg)\\
    &= \sum^M_{i = 1} w_i D_{f_\alpha}(x_i|| y_i).
\end{align}
We can see that for each particle ($\overline{x_i}(s), i\in [1,M]$), we can derive
\begin{align}
    \overline{x_i}(s) &= \argmin_{x_i(s)} \E_{a \sim \Bar{\pi}} [ D_{f_\alpha}(x_i(s)|| x_i(s,a))] \\
    \Longrightarrow \overline{x_i}(s) &= (\E_{a \sim \bar{\pi}} [x_i(s,a)^{p}])^{1/p},
\end{align}
with $p = 1 - \alpha$.
\end{proof}
\subsection{D. Convergence of Wasserstein Non-stationary multi-armed bandits}
We note that in an MCTS tree, each node is considered a non-stationary multi-armed bandit where the average mean drifts due to the given action selection strategy. Therefore, we first study the convergence of Wasserstein non-stationary multi-armed bandits where the action selection is Thompson sampling, with the power mean backup operator at the root node. Detailed descriptions of the Wasserstein Non-stationary multi-armed bandits settings can be found in the main article in the Theoretical Analysis section. 

We briefly summarize the theoretical results below. Theorem~\ref{theorem1} is about the upper bound on the expectation of the number of suboptimal arms playing, following the corresponding Theorem 4.2 in~\cite{jin2022finite}. Theorem~\ref{theorem2} is about the bias of the expected value of the power mean backup operator, which follows the result as Theorem 3 in power-UCT~\cite{dam2019generalized}. Theorem~\ref{theorem3} deals with the polynomial concentration of the power mean backup operator around the optimal mean at the root node of the non-stationary Wasserstein problem for multi-armed bandits. This theorem plays an important role in deriving the polynomial convergence of the choice of the optimal action at the root node in the Wasserstein MCTS tree, described in the next section. 

Let us define the event $E_{k,\epsilon}(t) = \{\theta_k(t) \leq \mu^* - \epsilon\}$ for all $k \in [K], \epsilon > 0, \theta_k(t)$ is sampled from $\mathcal{N}(\mu_k, V_k/T_k(n))$ at timestep $t$. Let us consider the decomposition

\begin{flalign}
    \E [T_k(T)] &= 1 + \E \Big[ \sum^{T}_{t = K + 1} \1 \{ A_t = a_k, E_{k,\epsilon}(t)\} 
    + \sum^{T}_{t = K + 1} \1 \{ A_t=a_k, E^c_{k,\epsilon}(t) \}\Big]\nonumber \\
                &= 1 + \underbrace{\E \Big[ \sum^{T}_{t = K + 1} \1 \{ A_t = a_k, E_{k,\epsilon}(t)\} \Big]}_{A} + \underbrace{\E \Big[ \sum^{T}_{t = K + 1} \1 \{ A_t = a_k, E^c_{k,\epsilon}(t)\} \Big]}_{B}.\label{expected_visitation}
\end{flalign}
Here $E^c $ is the complement of an event $E$, $\epsilon > 8\sqrt{V/T}$ is an arbitrary constant. 

\textbf{Bounding Term A:} Let's define 
\begin{flalign}
    \alpha_s = \sup_{x \in [0, \mu^{*} - \epsilon) } \Big\{ \text{KL}(\mu^{*} - \epsilon - x, \mu^{*}) \leq 4\log(\frac{T}{s})/s \label{alpha_s} \Big\}.
\end{flalign}
\begin{manuallemma}{1}\label{lemma1}
Let $M = \lceil 16V \log (T\epsilon^2/V)/\epsilon^2\rceil$, and $\alpha_s$ be the same as defined in (\ref{alpha_s}) then
\begin{flalign}
    \E \Big[ \sum^T_{t = K+1} \1 \{ A_t = a_k, E_{k,\epsilon}(t) \}\Big] \leq \sum^M_{s=1}\E \Big[ \Big( \frac{1}{1 - F^*_{s}(\mu^* - \epsilon)} - 1\Big).\1 \{ \overline{\mu}^{*}_s \in ( \mu^* - \epsilon - \alpha_s, 1]\} \Big] + \circleddash\Big(\frac{V}{\epsilon^2}\Big),
\end{flalign}
where $F^*_{s}$ is the CDF of Gaussian with mean $\overline{\mu}^{*}_s$, $\overline{\mu}^{*}_s$ is the average reward of the optimal arm after $s$ visitations.
\end{manuallemma}

\begin{proof}
    Please refer to Lemma A.1~\cite{jin2022finite}.
\end{proof}

\begin{manuallemma}{2}\label{lemma2}
Let $M = \lceil 16V \log (T\epsilon^2/V)/\epsilon^2\rceil$. Then
\begin{flalign}
    \sum^M_{s=1} \E_{\overline{\mu}^*_{s}} \Big[ \Big( \frac{1}{1 - F^*_{s}(\mu^*-\epsilon)}\Big).\1 \{ \overline{\mu}^{*}_s \in ( \mu^* - \epsilon - \alpha_s, 1]\} \Big] = \Theta \Big( \frac{V\log(T\epsilon^2/V)}{\epsilon^2} \big).
\end{flalign}
\end{manuallemma}
\begin{proof}
    Please refer to Lemma A.2~\cite{jin2022finite}
\end{proof}
\textbf{Bounding Term B:}
\begin{manuallemma}{3}\label{lemma3}
Let $N = \min \{\frac{1}{1-\frac{\text{KL}(\mu_k + \rho_k, \mu^*-\epsilon)}{\log (T\epsilon^2/V)}}, 2\} $. For any $\rho_k, \epsilon > 0$ that satisfies $\epsilon + \rho_k < \Delta_i$, then

\begin{flalign}
    \E \Big[ \sum^T_{t = K+1} \1\{ A_t=k,E^c_{k,\epsilon}(t) \}  \Big] \leq 1 + \frac{2V}{\rho^2_k} + \frac{V}{\epsilon^2} + \frac{N \log(T\epsilon^2/V)}{\text{KL} (\mu_k + \rho_k, \mu^*-\epsilon)}.
\end{flalign}
\end{manuallemma}
\begin{proof}
    Please refer to Lemma C.1~\cite{jin2022finite}.
\end{proof}


\noindent From Assumption 1, we derive the upper bound for the expectation of the number of plays of a suboptimal arm.
\begin{manualtheorem}{1}
Consider Thompson Sampling strategy (using power mean estimator) applied to a non-stationary problem where the pay-off sequence satisfies Assumption 1. Fix $\epsilon \geq 0$. Let $T_k(n)$ denote the number of plays of arm $k$. Then if $k$ is the index of a suboptimal arm, then each sub-optimal arm $k$ is played in expectation at most
\begin{flalign}
\E[T_k(n)] \leq \Theta \Bigg( 1 + \frac{V\log (n\Delta_k^2/V)}{\Delta_k^2} \Bigg).
\end{flalign}
\end{manualtheorem}
\begin{proof}
The proof of Theorem~\ref{theorem1} closely follows Theorem 4.2(\cite{jin2022finite}) by observing results from Lemma~\ref{lemma1},~\ref{lemma2},~\ref{lemma3}. 
From(~\ref{expected_visitation}), putting all Lemma~\ref{lemma1},~\ref{lemma2},~\ref{lemma3}, we have
\begin{flalign}
    \E [T_k(n)] = \Theta \Big( 1 + \frac{V \log (n\epsilon^2/V)}{(\Delta_k - \epsilon - \rho_k)^2} + \frac{V}{\rho^2_k} + \frac{V\log(n\epsilon^2/V)}{\epsilon^2} \Big).
\end{flalign}
Set $\epsilon = \rho_k = \Delta_k/4$, we derive
\begin{flalign}
    \E [T_k(n)] \leq \Theta \Big( 1 + \frac{V\log(n\Delta^2_k/V)}{\Delta^2_k} \Big).
\end{flalign}
\end{proof}
\begin{manualtheorem}{2}
Under Assumption 1, with $\triangle = \max\{\triangle_k\}, k \in [K]$ the following holds
\begin{flalign}
\big| \E\big[ \overline{X}_n(p) \big]  - \mu^{*} \big| &\leq |\delta^*_n| + \Theta \Big(\frac{(K-1)(1+V\log(n\triangle^2/V))}{\triangle^2n} \Big)^{\frac{1}{p}}. \nonumber
\end{flalign}
\end{manualtheorem} 
\begin{proof}
We have
\begin{flalign}
\left|\E\left[ \overline{X}_n(p) \right] - \mu^{*}\right| \leq |\delta^*_n| + \E\left[ \overline{X}_n(p) \right] - \mu_{n}^{*} = |\delta^*_n| + \left| \E\left[ \left(\sum^{K}_{i=1} \frac{T_k(n)}{n} \overline{X}_{i, T_k(n)}^p\right)^{\frac{1}{p}} \right] - \mu_{n}^{*}\right|.
\end{flalign}
In the proof, we will make use of the following inequalities:
\begin{flalign}
\label{reward_01} &0 \leq X_i \leq 1, \\
\label{13} &x^{\frac{1}{p}} \leq y^{\frac{1}{p}} \space \text{ when 0 } \leq {x} \leq { y }, \\
\label{14} &(x + y)^m \leq x^m + y^m \space (0 \leq m \leq 1), \\
\label{15} &\E[f(X)] \leq f(\E[X]) \textit{ (f(X)} \text{ is concave)}.
\end{flalign}
With $i^*$ being the index of the optimal arm, we can derive an upper bound on the difference between the value backup and the true average reward
\begin{flalign}
&\E\left[ \left(\sum^{K}_{k=1} \frac{T_k(n)}{n} \overline{X}_{k, T_k(n)}^p\right)^{\frac{1}{p}} \right] - \mu_n^{*} \leq \E\left[ \left( \left(\sum^{K}_{k=1;k\neq k^{*}} \frac{T_k(n)}{n} \right) + \overline{X}_{k^*, T_k^*(n)}^p\right)^{\frac{1}{p}} \right] - \mu_n^{*} (\text{see } (\ref{reward_01})) \nonumber\\
&\leq \E\left[ \left(\sum^{K}_{k=1;k\neq k^{*}} \frac{T_k(n)}{n} \right)^{\frac{1}{p}} + \overline{X}_{k^*, T_k^*(n)} \right] - \mu_n^{*} (\text{see } (\ref{14}))\nonumber\\
&= \E\left[ \left(\sum^{K}_{k=1;k\neq k^{*}} \frac{T_k(n)}{n} \right)^{\frac{1}{p}} \right] + \E \left[\overline{X}_{k^*, T_k^*(n)}\right]  - \mu_n^{*} = \E\left[ \left(\sum^{K}_{k=1;k\neq k^{*}} \frac{T_k(n)}{n} \right)^{\frac{1}{p}} \right] \leq \left(\sum^{K}_{k=1;k\neq k^{*}} \E \left[\frac{T_k(n)}{n}\right] \right)^{\frac{1}{p}} (\text{see } (\ref{15})) \nonumber\\
\label{16} &\leq \Theta \Big(\frac{(K-1)(1+V\log(n\triangle^2/V))}{\triangle^2n} \Big)^{\frac{1}{p}} (\text{Theorem~\ref{theorem1} \& } (\ref{13})).
\end{flalign}
According to Lemma 1~\cite{dam2019generalized}, it holds that
$\E\left[\overline{X}_n(p)\right] \geq \E\left[\overline{X}_n(1)\right].$
for $p \geq 1$.

We have $\overline{X}_n(1) = \sum^{K}_{k=1} \left(\frac{T_{k}(n)}{n}\right) \overline{X}_{k, T_k(n)}$. 
We can observe that 
\begin{flalign}
&\big| \E [\overline{X}_n(1)] - \mu_{n}^{*}) \big|\leq \bigg| \E \bigg [\sum^{K}_{k=1} \left(\frac{T_{k}(n)}{n}\right) \overline{X}_{k, T_k(n)}\bigg] - \mu_{n}^{*})\bigg| \leq \underbrace{\bigg| \E \bigg [\sum^{K}_{k \neq k_{*}} \left(\frac{T_{k}(n)}{n}\right) \overline{X}_{k, T_k(n)}\bigg] \bigg|}_{\text{$B_1$}}\nonumber\\ 
&+ \underbrace{\bigg|\E\bigg[ \frac{T_{k^*}(n)}{n}\overline{X}_{k^*, T_k(n)} \bigg] - \mu_{n}^{*})\bigg|}_{\text{$B_2$}}\nonumber
\end{flalign}
Using results from Theorem~\ref{theorem1}, and the fact that $\overline{X}_{k, T_k(n)} \leq 1$, we can upper bound the first term as
\begin{flalign}
    B_1 &\leq \bigg| \E \bigg [\sum^{K}_{k \neq k_{*}} \left(\frac{T_{k}(n)}{n}\right)\bigg] \bigg|\leq \bigg| \E \bigg [\sum^{K}_{k \neq k_{*}} \left(\frac{T_{k}(n)}{n}\right)\bigg] \bigg|^{\frac{1}{p}} \leq \Theta (K-1)\Big(\frac{(1+V\log(n\triangle^2/V))}{\triangle^2n} \Big)^{\frac{1}{p}},
\end{flalign}
Here $\triangle = \max_{k\in[K]}\{\triangle_k\}$.

To bound the second term, we have 
\begin{flalign}
    B_2 &= \bigg |\frac{\E [ \sum_1^{T_{k^*}(n)} \overline{X}_{k^*, T_k(n)} - n \mu_{n}^{*})]}{n} \bigg|\\
    &= \bigg | \frac{\E [\sum^{n}_{i=T_{k^*}(n)+1} \mu_{n}^{*})] }{n}\bigg| \leq \bigg | \frac{\E[ n -  T_{k^*}(n)]}{n}\bigg| \\
    &=  \bigg|\sum_{k \neq k^*}\frac{\E [ T_k(n) ]}{n} \bigg| 
    \leq \bigg| \E \bigg [\sum^{K}_{k \neq k_{*}} \left(\frac{T_{k}(n)}{n}\right)\bigg] \bigg|^{\frac{1}{p}} \leq \Theta (K-1)\Big(\frac{(1+V\log(n\triangle^2/V))}{\triangle^2n} \Big)^{\frac{1}{p}},
\end{flalign}
So that we have 
\begin{flalign}
\big| \E [\overline{X}_n(1)] - \mu_{n}^{*}) \big| &\leq \Theta (K-1)\Big(\frac{(1+V\log(n\triangle^2/V))}{\triangle^2n} \Big)^{\frac{1}{p}}. \nonumber
\end{flalign}
That leads to 
\begin{flalign}
\E [\overline{X}_n(p)] - \mu_{n}^{*} \geq \E [\overline{X}_n(1)] - \mu_{n}^{*} \geq -\Theta (K-1)\Big(\frac{(1+V\log(n\triangle^2/V))}{\triangle^2n} \Big)^{\frac{1}{p}}. \label{17}
\end{flalign}

Combining (\ref{16}) and (\ref{17}) concludes our prove
\begin{flalign}
\big| \E\big[ \overline{X}_n(p) \big]  - \mu^{*} \big| &\leq |\delta^*_n| + \Theta (K-1)\Big(\frac{(1+V\log(n\triangle^2/V))}{\triangle^2n} \Big)^{\frac{1}{p}}. \nonumber
\end{flalign}
\end{proof}

\begin{manuallemma}{4}\label{lemma4}
    Define $A(t) = 1 + \frac{V\log(t\triangle^2/V)}{\triangle^2} $, where $\triangle = \max_{k \in [K]} \{\triangle_k\}$. We further denote $N_p$ by the first time such that $t \geq A(t)$ as
    \begin{flalign}
        N_p = \min\{t\geq 1: t \geq A(t)\}.
    \end{flalign}
    For any $n \geq N_p, \epsilon > 0$ and $x \geq 0, $ let $\epsilon_0 = \frac{n\epsilon}{x} + \frac{n(K-1)}{x}(\frac{A(n)+\log x}{n})^{\frac{1}{p}}, \exists \alpha > 0$. Then,
    \begin{flalign}
        &\Pr \bigg(\overline{X}_n(p) - \mu^{*}  \geq \frac{\epsilon_0 x}{n} \bigg ) \leq \exp\{-n\epsilon^2/2V\} + C (n x)^{-\alpha} \label{lemma_4_first_eq}\\ 
        &\Pr \bigg(\overline{X}_n(p) - \mu^{*}  \leq -\frac{\epsilon_0 x}{n} \bigg ) \leq \exp\{-n\epsilon^2/2V\} + C (n x)^{-\alpha} \label{lemma_4_second_eq}
    \end{flalign}
\end{manuallemma}
\begin{proof}
    We first prove~(\ref{lemma_4_first_eq}), and the other direction follows the same steps.
    (\ref{lemma_4_second_eq}) follows the same steps.
    We have $\overline{X}_n(p) = \left(\sum^{K}_{k=1} \frac{T_{k}(n)}{n} \overline{X}^{p}_{k,n}\right)^{1/p}$. 
    Therefore,
    \begin{flalign}
        &\overline{X}_n(p) - \mu^{*} = \overline{X}_{k_{*},n} -\mu^{*} + \left(\sum^{K}_{k=1} \frac{T_{k}(n)}{n} \overline{X}^{p}_{k,n}\right)^{1/p} - \overline{X}_{k_{*},n} \leq 
        \overline{X}_{k_{*},n} -\mu^{*} + \left(\sum^{K}_{k=1} \frac{T_{k}(n)}{n} + \overline{X}^p_{k_{*},n} \right)^{1/p} - \overline{X}_{k_{*},n}\\
        &\leq \overline{X}_{k_{*},n} -\mu^{*} + \left(\sum^{K}_{k=1,k \neq k_*} \frac{T_{k}(n)}{n}\right)^{1/p} + \overline{X}_{k_{*},n} - \overline{X}_{k_{*},n} \leq \overline{X}_{k_{*},n} -\mu^{*} + \sum^{K}_{k=1,k\neq k^*} \left(\frac{T_{k}(n)}{n}\right)^{1/p}.
    \end{flalign}
    Then we have 
    \begin{flalign}
        &\Pr \bigg( \overline{X}_n(p) - \mu^{*}  \geq \frac{\epsilon_0 x}{n} \bigg) \leq \Pr \bigg( \overline{X}_{k_{*},n} - \mu^{*} + \sum^{K}_{k=1,k\neq k^*} (\frac{T_{k}(n)}{n})^{\frac{1}{p}} \geq \frac{\epsilon_0 x}{n}\bigg)\\
        &\leq \underbrace{\Pr \bigg( \overline{X}_{k_{*},n} - \mu^{*} \geq \epsilon \bigg)}_{\text{$C_1$}} + \underbrace{\sum_{k \neq k^{*}} \Pr \bigg( \frac{T_{k}(n)}{n} \geq \frac{A(n)+\log x}{n} \bigg)}_{\text{$C_2$}}. \label{C1_C2}
    \end{flalign}
From the Assumption~\ref{assumpt1}, we can derive $C_1 \leq \exp\{-n\epsilon^2/2V_{k_*}\} \leq \exp\{-n\epsilon^2/2V\}$.\\
We study two following events (with $0<u<n$)
\begin{flalign}
    &E_1 = \{ \text{for each integer } t \in [u,n], \text{ we have } \overline{\theta}_k(t) < \mu^*-\epsilon\},\\
    &E_2 = \{ \text{for each integer } t \in [u,n], \text{ we have } \overline{\theta}_*(t) > \mu^*-\epsilon\}.
\end{flalign}   
We see that $E_1 \cap E_2 \implies T_k(n) < u$, so that $\{T_k(n) \geq u\} \subset \{E^c_1\} \cup \{E^c_2\}$. Therefore, 
\begin{flalign}
    \Pr(T_k(n) \geq u) \leq \Pr(\exists u \leq t_0 \leq n, \overline{\theta}_k(t_0) \geq \mu^*-\epsilon_0) + \Pr(\exists u \leq t_0 \leq n, \overline{\theta}_*(t_0) \leq \mu^*-\epsilon_0).
\end{flalign}
We choose $\epsilon_0$ so that $\mu^* - \epsilon_0 > \mu_k$, then 
\begin{flalign}
    &\Pr(\exists u \leq t_0 \leq n, \overline{\theta}_k(t_0) \geq \mu^*-\epsilon_0) = \frac{1}{2}\exp\{-ub_n \text{KL}(\mu_k,\mu^*-\epsilon_0)\} ~(\text{Proposition 4.1~\cite{jin2022finite}})\\
    &= \frac{1}{2}\exp\{-u\frac{1}{2}\text{KL}(\mu_k,\mu^*-\epsilon_0)\} \text{(Because we consider Gaussians, $b_n=1/2$)}.
\end{flalign}
\noindent Furthermore, $\text{KL}(\mu_k, \mu^*-\epsilon_0) = (\mu_k-\mu^*+\epsilon_0)^2/2V$. Set $u = A(n)+ \log x$. Thus, $\exists C_1 > 0, \alpha_1 > 0$ that for any $n \geq N_p$
\begin{flalign}
    &\Pr(\exists u \leq t_0 \leq n, \overline{\theta}_k(t_0) \geq \mu^*-\epsilon_0) = \frac{1}{2}\exp\{\frac{-(A(n) + \log x)(\mu_k-\mu^*+\epsilon_0)^2}{4V}\} \leq C_1 (n x)^{-\alpha_1}.
\end{flalign}
Similarly, $\exists C_2>0, \alpha_2>0$ that for any $n \geq N_p$
\begin{flalign}
    &\Pr(\exists u \leq t_0 \leq n, \overline{\theta}_*(t_0) \leq \mu^*-\epsilon_0) = \frac{1}{2}\exp\{-ub_n \text{KL}(\mu^*,\mu^*-\epsilon_0)\} ~(\text{Proposition 4.1~\cite{jin2022finite}})\\
    &= \frac{1}{2}\exp\{-u\frac{1}{2}\text{KL}(\mu^*,\mu^*-\epsilon_0)\} 
    =\frac{1}{2}\exp\{-\frac{(A(n) + \log x)\epsilon^2_0)}{4V}\}
    \leq C_2 (n x)^{-\alpha_2}.
\end{flalign}
Therefore, with $\exists C >0, \alpha > 0$, that for any $n \geq N_p$
    \begin{flalign}
        &\Pr \bigg(\overline{X}_n(p) - \mu^{*}  \geq \frac{\epsilon_0 x}{n} \bigg ) \leq  \exp\{-n\epsilon^2/2V\} + C (n x)^{-\alpha}.
    \end{flalign}
\noindent From Lemma~\ref{lemma4}, we derive the following result
\end{proof}
\begin{manualtheorem}{3}
    $\exists$ constant $C_0>0, \alpha > 0, \beta > 0$ with $\alpha=2\beta$ that for any $\epsilon > 0, \exists N^{'}_p>0$, that for any $n \geq N^{'}_p$ we can derive
    \begin{flalign}
        &\Pr \bigg(\overline{X}_n(p) - \mu^{*} \geq \epsilon \bigg ) \leq Cn^{-\alpha}\epsilon^{-\beta},\\
        &\Pr \bigg(\overline{X}_n(p) - \mu^{*} \leq -\epsilon \bigg ) \leq Cn^{-\alpha}\epsilon^{-\beta}.
    \end{flalign}
\end{manualtheorem}
\begin{proof}  
We will use results from Lemma~\ref{lemma4} to derive the proof for Theorem~\ref{theorem3}.
We will prove the first direction of inequality. The other direction follows the same steps. We want to upper bound 
\begin{flalign}
    D = \Pr \bigg(\overline{X}_n(p) - \mu^{*} \geq \epsilon \bigg ).
\end{flalign}
Due to Lemma~\ref{lemma4}, we can find $C_0>0, \alpha_0>0, N_p>0$ that with $n \geq N_p$, we have
\begin{flalign}
    \Pr \bigg(\overline{X}_n(p) - \mu^{*} \geq \frac{\epsilon_0 x}{n} \bigg ) \leq \exp\{-n\epsilon^2/2V\} + C_0 (n x)^{-\alpha_0} \leq C_1 (n x)^{-\alpha_0} (\exists C_1>0 \text{ big enough}).
\end{flalign}
We set $\epsilon = \frac{\epsilon_0 x}{n}.$
We have $x = \frac{n\epsilon}{\epsilon_0}$
Then $\exists C_0 =\frac{C_1}{\epsilon^{-\alpha}_0}, \alpha =2 \alpha_0, \beta =\alpha_0$, that we can derive
\begin{flalign}
        &\Pr \bigg(\overline{X}_n(p) - \mu^{*}  \geq \epsilon \bigg ) \leq  C_0n^{-\alpha}\epsilon^{-\beta}.
\end{flalign}
The other direction follows the same steps. That concludes the proof.
\end{proof}
\subsection{E. Convergence of Wasserstein Monte-Carlo tree search}
Based on the results of the described nonstationary multi-armed bandit problem, we derive theoretical results for W-MCTS-TS in an MCTS tree. 

We derive Proposition~\ref{proposition3}, which shows the polynomial concentration of the estimated mean of the Q-value function at the root node. In Proposition~\ref{proposition3}, we also show that the estimated mean of the V-value function at the root node converges polynomially to the optimal mean.
Based on the results of Proposition~\ref{proposition3}, we estimate in Theorem~\ref{theorem4} that the probability of not choosing the optimal action at the root node decreases polynomially towards zero. 
Finally, in Theorem~\ref{theorem5}, we show the results about the bias of the expected payoff of the power mean backup at the root node.\\
\begin{manuallemma}{5}\label{lemma5}
    Consider real-valued random variables $X_i, Y_i$ for $i\geq 1$ such that $Xs$ are independent and identically distributed, taking values in [0, 1], $\E[X_i] = \mu_X$, $Xs$ are independent of $Ys$, and $Ys$ satisfy\\
    
    (1). For $n\geq 1$, with notation $\overline{Y}_n = \frac{1}{n}(\sum^n_{i=1}) Y_i$,
    \begin{flalign}
        \lim_{n\rightarrow\infty} \E[\overline{Y}_n] = \mu_Y.
    \end{flalign}

    (2). There exists constants $C_0>0, \alpha> 0, \beta > 0, N_p>0$ that for any $\epsilon>0$, $n \geq N_p$, we have
    \begin{flalign}
        &\Pr \bigg(\overline{Y}_n - \mu_Y \geq \epsilon \bigg ) \leq C_0 n^{-\alpha} \epsilon^{-\beta},\\
        &\Pr \bigg(\overline{Y}_n - \mu_Y \leq -\epsilon \bigg ) \leq C_0 n^{-\alpha} \epsilon^{-\beta}.
    \end{flalign}

Let $Z_i = X_i + \gamma Y_i$, $\gamma > 0$. Then $Zs$ satisfies

    (1). For $n\geq 1$, with notation $\overline{Z}_n = \frac{1}{n}(\sum^n_{i=1}) Z_i$,
    \begin{flalign}
        \lim_{n\rightarrow\infty} \E[\overline{Z}_n] = \mu_X + \gamma\mu_Y.
    \end{flalign}

    (2). There exists constants $C_1>0, N_p>0$ that for any $\epsilon>0$, $n \geq N_p$, we have
    \begin{flalign}
        &\Pr \bigg(\overline{Z}_n - (\mu_X + \gamma\mu_Y) \geq \epsilon \bigg ) \leq C_1 n^{-\alpha} \epsilon^{-\beta},\\
        &\Pr \bigg(\overline{Z}_n - (\mu_X + \gamma\mu_Y) \leq -\epsilon \bigg ) \leq C_1 n^{-\alpha} \epsilon^{-\beta}.
    \end{flalign}
\end{manuallemma}
\begin{proof}
    
    (1) Due to the linearity of the expectation, we can derive $lim_{n\rightarrow\infty} \E[\overline{Z}_n] = \mu_X + \gamma\mu_Y.$
    
    (2) Due to the Hoeffding’s inequality, we can have
    There exists constants $C_0>0, N_p>0$ that for any $\epsilon>0$, $n \geq N_p$, we have
    \begin{flalign}
        &\Pr \bigg(\overline{X}_n - \mu_X \geq \epsilon \bigg ) \leq \exp\{-2n\epsilon^2\},\\
        &\Pr \bigg(\overline{X}_n - \mu_Y \leq -\epsilon \bigg ) \leq \exp\{-2n\epsilon^2\}.
    \end{flalign}
    Therefore,
    \begin{flalign}
        &\Pr \bigg(\overline{Z}_n - (\mu_X + \gamma\mu_Y) \geq \epsilon \bigg ) \leq \Pr \bigg( \overline{X}_n - \mu_X \geq \frac{\epsilon}{2} \bigg) + \Pr \bigg( \gamma\overline{Y_n} -\gamma \mu_Y \geq \frac{\epsilon}{2} \bigg)\\
        &\leq \exp\{-2n(\epsilon/2)^2\} + C_0 n^{-\alpha} (\epsilon/(2\gamma))^{-\beta}.
    \end{flalign}
    Therefore, $\exists C_1>0, N_p > 0$ that for any $n \geq N_p$, $ \Pr \bigg(\overline{Z}_n - (\mu_X + \gamma\mu_Y) \geq \epsilon \bigg ) \leq C_1 n^{-\alpha} \epsilon^{-\beta}.$
    The other direction follows the same steps. That concludes the proof.
\end{proof}
At any node of state $s^{(h)}$ at depth $h$ in the tree, the mean of the Q value function, and the mean value of the optimal value function are defined as
\begin{flalign}
    Q_m^{(h)}(s^{(h)},a) &= \E_{\mathcal{P}}[r(s^{(h)},a)] + \gamma V_m^{(h+1)} (s^{(h+1)}),\nonumber\\
    V_m^{(h)}(s^{(h)}) &= \argmax_a Q_m^{(h)}(s^{(h)},a),\nonumber
\end{flalign}
with $h = [H-1,...,1,0],~V^{(H)}(s^{(H)}) $ is the value return from rollouts at state $s^{(H)}$, $\mathcal{P}$ is the stochastic transition dynamic from environment. Let us denote $a_{k^{*}}$ as the optimal action at the root node. 

\begin{manualproposition}{3}\label{proposition3}
When we apply the W-MCTS algorithm to an MCTS tree of depth $(H)$, we have

(1). For $\epsilon>0$, $\exists C_0 > 0, \alpha> 0, \beta >0$ as constant that
\begin{flalign}
        &\Pr \bigg( \overline{Q}_m^{(0)}(s^{(0)},a_k,n) - Q_m^{(0)}(s^{(0)},{a_{k}}) \geq \epsilon \bigg ) \leq  C_0 n^{-\alpha} \epsilon^{-\beta}.\\
        &\Pr \bigg( \overline{Q}_m^{(0)}(s^{(0)},a_k,n) - Q_m^{(0)}(s^{(0)},{a_{k}}) \leq -\epsilon \bigg ) \leq  C_0 n^{-\alpha} \epsilon^{-\beta}.
\end{flalign}

(2). Let us denote $T^{(0)}_{s^{(0)},a_k}(n)$ as the number of plays of action $a_k$ at state $s^{(0)}$ at timestep $n$. Then if $a_k$ is a sub-optimal action, $a_{k^*}$ is the optimal action. With $\triangle_k = Q_m^{(0)}(s^{(0)},{a_{k^*}}) - Q_m^{(0)}(s^{(0)},{a_{k}})$, we have
\begin{flalign}
\E[T^{(0)}_{s^{(0)},a_k}(n)] \leq \Theta \Big( 1 + \frac{V\log(n\Delta^2_k/V)}{\Delta^2_k} \Big).
\end{flalign}

(3). $\exists$ constant $C_0>0, \alpha>0, \beta>0$ that for any $\epsilon > 0, n \geq 1$, we can derive
    \begin{flalign}
        &\Pr \bigg(Q_m^{(0)}(s^{(0)},{a_{k^*}}) - \overline{V}_m^{(0)}(s^{(0)},n) \geq \epsilon \bigg ) \leq C_0 n^{-\alpha} \epsilon^{-\beta}.\\ 
        &\Pr \bigg(Q_m^{(0)}(s^{(0)},{a_{k^*}}) - \overline{V}_m^{(0)}(s^{(0)},n) \leq -\epsilon \bigg ) \leq C_0 n^{-\alpha} \epsilon^{-\beta}.
    \end{flalign}
\end{manualproposition}
\begin{proof}
    We will prove this by induction. 
    If the tree only has depth $(1)$. (1) is a direct result of Lemma~\ref{lemma5} with $X_i$ is the intermediate reward $r_i(s^{(0)},a_k)$, $Y_i$ is the estimated mean value at state $s^{(1)}$ ($s^{(1)}\sim \tau(s^{(0)},a_k)$, where $\tau(s,a_k)$ is the probability transition dynamic of taking action $a_k$ at state $s^{(0)}$), $\overline{V}^{(1)}_m(s^{(1)})$ is the deterministic initial Value function from rollouts.
    
    (2) is correct due to Theorem~\ref{theorem1}.

    (3) direct results from Theorem~\ref{theorem3}.\\

Let's assume that the theorem is correct with tree of depth $(h)$.
    Now let's consider the tree with depth $(h+1)$. When we take one action at the root node, it comes to a subtree with depth $(h)$. We have $s^{(1)} = s^{(0)} \omicron a_k$ 
    Then we can derive
    \begin{flalign}
        &Q_m^{(0)}(s^{(0)},a_k) = \E_{s^{(1)}, s^{(0)},a_k} [r(s^{(0)},a_k)] + \gamma V^{(1)}(s^{(1)}).
    \end{flalign}
    Furthermore, at the root node of a subtree with depth $(h)$, $\exists$ constant $C_0>0, N^{(1)}_p > 0$ that for any $\epsilon>0, n \geq N^{(1)}_p$, we have
    \begin{flalign}
        \Pr \Bigg( \overline{Q}_m^{(1)}(s^{(1)},a_k,n) - Q_m^{(1)}(s^{(1)},{a_{k}}) \geq \epsilon \Bigg) \leq C_0 n^{-\alpha} \epsilon^{-\beta}.
    \end{flalign}
For any sub-optimal action $a_k$
    \begin{flalign}
    \E[T^{(1)}_{s^{(1)},a_k}(n)] \leq \Theta \Big( 1 + \frac{V\log(n\Delta^2_k/V)}{\Delta^2_k} \Big).
    \end{flalign}
Also,  $\exists$ constants $C_0>0, N^{(1)}_p>0$ that for any $\epsilon > 0, n \geq N^{(1)}_p$, we can derive
    \begin{flalign}
        &\Pr \bigg(Q_m^{(1)}(s^{(1)},{a_{k^*}}) - \overline{V}_m^{(1)}(s^{(1)},n) \geq \epsilon \bigg ) \leq C_0 n^{-\alpha} \epsilon^{-\beta} \label{depth_h_first_eq},\\ 
        &\Pr \bigg(Q_m^{(1)}(s^{(1)},{a_{k^*}}) - \overline{V}_m^{(1)}(s^{(1)},n) \leq -\epsilon \bigg ) \leq C_0 n^{-\alpha} \epsilon^{-\beta} \label{depth_h_second_eq}.
    \end{flalign}
\noindent (1) follows as now at the root node, we have a direct result of Lemma~\ref{lemma5} with $X_i$ as the intermediate reward $r_i(s^{(0)},a_k)$, $Y_i$ is the estimated mean value at state $s^{(1)}$ ($s^{(1)}\sim \tau(s^{(0)},a_k)$, where $\tau(s^{(0)},a_k)$ is the probability transition dynamic of taking action $a_k$ at state $s^{(0)}$), and  the concentration condition of $Y_i$ according to (~\ref{depth_h_first_eq}),(\ref{depth_h_second_eq}), we have for $\epsilon>0$, $\exists C_0 > 0$ as constant that
\begin{flalign}
        &\Pr \bigg( \overline{Q}_m^{(0)}(s^{(0)},a_k,t) - Q_m^{(0)}(s^{(0)},{a_{k}}) \geq \epsilon \bigg ) \leq  C_0 n^{-\alpha} \epsilon^{-\beta}.\\
        &\Pr \bigg( \overline{Q}_m^{(0)}(s^{(0)},a_k,t) - Q_m^{(0)}(s^{(0)},{a_{k}}) \leq -\epsilon \bigg ) \leq C_0 n^{-\alpha} \epsilon^{-\beta}.
\end{flalign}

\noindent For (2), let us consider the Thompson Sampling for action selection of the W-MCTS algorithm at the root node $s^{(0)}$. Due to Assumption 1, each Q Value function at state action $(s^{(0)},a_k)$ is a Gaussian with mean value $Q_m^{(0)}(s^{(0)},a_k)$, and as a result of Theorem~\ref{theorem2}, we can get

\begin{flalign}
E[T^{(0)}_{s^{(0)},a_k}(t)] \leq  \Theta \Big( 1 + \frac{V\log(n\Delta^2_k/V)}{\Delta^2_k} \Big).
\end{flalign}
(3) follows the same steps as in Theorem~\ref{theorem3}.

\noindent By induction, (1), (2), (3) is correct with the tree of depth $(h+1)$, which concludes the proof.
\end{proof}

\begin{manualtheorem}{4}
Let $a_k$ be the action returned by W-MCTS at timestep n at the root node. Then for $\epsilon>0$, $\exists C> 0, \alpha > 0, N_p > 0$ as constants that for all $n \geq N_p$, we have
\begin{flalign}
        &\Pr \bigg( a_k \neq a_{k^*} \bigg ) \leq  C n^{-\alpha}.
\end{flalign}
\end{manualtheorem}
\begin{proof}
    At the root node $s^{(0)}$, let us define
    \begin{flalign}
    \overline{\mu}_{k}(t) &= \overline{Q}_m^{(0)}(s^{(0)}, a_k, t)\text{ as an empirical estimate of Q value function at  state $s^{(0)}$, action $a_k$ after t rounds}.\\
    \overline{\mu}_{k^{*}}(t) &= \overline{Q}_m^{(0)}(s^{(0)}, a_{k^{*}}, t)\text{ as an empirical estimate of Q value function at  state $s^{(0)}$, optimal action $a_{k^*}$ after t rounds}.\\
    \mu_{k^*} &= {Q}_m^{(0)}(s^{(0)}, a_{k^*})\text{ as the mean of Q value function at  state $s^{(0)}$, optimal action $a_{k^*}$ }.\\
    \mu_{k} &= {Q}_m^{(0)}(s^{(0)}, a_k)\text{ as the mean of Q value function at state $s^{(0)}$, action $a_k$}.\\
\end{flalign}
Without loss of generality, let us also define $a_{k^*}$ as a unique optimal action at the root node $s^{(0)}$.
We have 
\begin{flalign}
    \Pr \bigg(\exists k \in [K], \overline{\mu}_{k}(t) > \overline{\mu}_{k^{*}}(t) \bigg) \leq \sum_{k} \underbrace{\Pr \bigg(\overline{\mu}_{k}(t) > \overline{\mu}_{k^{*}}(t) \bigg)}_{\text{$K$}}.
\end{flalign}
We will upper bound $K$. We see that we can find $\xi > 0, \nu > 0$ that $\xi\mu_k + \nu \mu_{k^{*}} = \mu_{k^{*}} - \mu_k$. Therefore,
\begin{flalign}
    K &= \Pr \bigg(\overline{\mu}_{k}(t) - \mu_k + \mu_{k^{*}} - \overline{\mu}_{k^{*}}(t) > \mu_{k^*} -\mu_{k} \bigg) \leq \Pr \bigg( \overline{\mu}_{k}(t) - \mu_k > \alpha \mu_k \bigg) + \Pr \bigg ( \mu_{k^{*}} - \overline{\mu}_{k^{*}}(t) > \beta \mu_{k^{*}} \bigg) \\
    &\leq C_0 n^{-\alpha} (\xi \mu_k)^{-\beta} + C_1 n^{-\alpha} (\nu \mu_{k^*})^{-\beta}.
\end{flalign}
Therefore, $\exists C > 0$, that
\begin{flalign}
    &\Pr \bigg( a_k \neq a_{k^*} \bigg ) = \Pr \bigg(\exists k \in [K], \overline{\mu}_{k}(t) > \overline{\mu}_{k^{*}}(t) \bigg) \leq \sum_{k} C_0 n^{-\alpha} (\xi \mu_k)^{-\beta} + C_1 n^{-\alpha} (\nu \mu_{k^*})^{-\beta}\\
    &\leq C n^{-\alpha}.
\end{flalign}
That concludes the proof.
\end{proof}

\begin{manualtheorem} {5}
We have at the root node $s^{(0)}$, $\exists N_0>0, \triangle = \max_{k\in[K]}\{\triangle_k\}$ so that the expected payoff satisfies
\begin{flalign}
\big| \E [\overline{V}_m^{(0)}(s^{(0)},n)] - Q_m^{(0)}(s^{(0)},{a_{k^*}}) \big| &\leq \Theta \Bigg( \frac{2(K-1)(1 + \frac{V\log(n\triangle^2/V)}{\triangle^2})}{n} \Bigg). \nonumber
\end{flalign}
\end{manualtheorem}
\begin{proof}
In W-MCTS, the value of each node is used for backup as $\overline{V}_m^{(0)}(s^{(0)},n) = \sum^{K}_{k=1} \left(\frac{T^{(0)}_{s^{(0)},a_k}(n)}{n}\right) \overline{Q}_m^{(0)}(s^{(0)},{a_k}, T^{(0)}_{s^{(0)},a_k}(n))$. 
Without the loss of generality, we assume that there is a unique optimal Q value function $Q_m^{(0)}(s^{(0)}, a_{k^*})$, we can observe that 
\begin{flalign}
&\big| \E [\overline{V}_m^{(0)}(s^{(0)},n)] - Q_m^{(0)}(s^{(0)},{a_{k^*}}) \big|\\
&\leq \bigg| \E \bigg [\sum^{K}_{k=1} \left(\frac{T^{(0)}_{s^{(0)},a_k}(n)}{n}\right) \overline{Q}_m^{(0)}(s^{(0)},{a_k},T^{(0)}_{s^{(0)},a_k}(n))\bigg] - Q_m^{(0)}(s^{(0)},{a_{k^*}})\bigg|\nonumber \\
&\leq \underbrace{\bigg| \E \bigg [\sum^{K}_{k \neq k_{*}} \left(\frac{T^{(0)}_{s^{(0)},a_k}(n)}{n}\right) \overline{Q}_m^{(0)}(s^{(0)},{a_k},T^{(0)}_{s^{(0)},a_k}(n))\bigg] \bigg|}_{\text{$B_1$}}\nonumber\\ 
&+ \underbrace{\bigg|\E\bigg[ \frac{T^{(0)}_{s^{(0)},a_{k^*}}(n)}{n}\overline{Q}_m^{(0)}(s^{(0)},{a_{k_{*}}},T^{(0)}_{s^{(0)},a_{k^*}}(n)) \bigg] - Q_m^{(0)}(s^{(0)},{a_{k^*}})\bigg|}_{\text{$B_2$}}.\nonumber
\end{flalign}
Using results from Proposition~\ref{proposition3}, and the fact that $\overline{Q}_m^{(0)}(s^{(0)},a_k,t) \leq 1$ (because all the intermediate rewards at timestep $t$~ $r_t(s^{(0)},a_k) \in [0,1]$), we can upper bound the first term as
\begin{flalign}
    B_1 \leq \Theta \Bigg( \frac{(K-1)(1 + \frac{V\log(n\triangle^2/V)}{\triangle^2})}{n} \Bigg),
\end{flalign}
Here $\triangle = \max_{k\in[K]}\{\triangle_k\}$.

\noindent To bound the second term, we have 
\begin{flalign}
    B_2 &= \bigg |\frac{\E [ \sum_1^{T^{(0)}_{s^{(0)},a_{k^*}}(n)} \overline{Q}_m^{(0)}(s^{(0)},{a_{k_{*}}},T^{(0)}_{s^{(0)},a_{k^*}}(n)) - n Q_m^{(0)}(s^{(0)},{a_{k^*}})]}{n} \bigg|\\
    &= \bigg | \frac{\E [\sum^{n}_{i=T^{(0)}_{s^{(0)},a_{k^*}}(n)+1} Q_m^{(0)}(s^{(0)},{a_{k^*}})] }{n}\bigg| \leq \bigg | \frac{\E[ n -  T^{(0)}_{s^{(0)},a_{k^*}}(n)]}{n}\bigg| \\
    &=  \bigg|\sum\E_{k \neq k^*} [ T^{(0)}_{s^{(0)},a_{k}}(n) ] \bigg| 
    \leq  \Theta \Bigg( \frac{(K-1)(1 + \frac{V\log(n\triangle^2/V)}{\triangle^2})}{n} \Bigg).
\end{flalign}
So that we have 
\begin{flalign}
\big| \E [\overline{V}_m^{(0)}(s^{(0)},n)] - Q_m^{(0)}(s^{(0)},{a_{k^*}}) \big| &\leq \Theta \Bigg( \frac{2(K-1)(1 + \frac{V\log(n\triangle^2/V)}{\triangle^2})}{n} \Bigg). \nonumber
\end{flalign}
That concludes the proof.
\end{proof}

\end{document}